\documentclass[lettersize,journal]{IEEEtran}

\usepackage{graphicx}
\usepackage{amsmath}
\usepackage{amsfonts}
\usepackage{booktabs}
\usepackage{multirow}
\usepackage{threeparttable}
\usepackage{amssymb}
\usepackage{pifont}
\usepackage{bm}

\usepackage{algorithmic}
\usepackage{algorithm}
\usepackage{array}
\usepackage[caption=false,font=normalsize,labelfont=sf,textfont=sf]{subfig}
\usepackage{textcomp}
\usepackage{stfloats}
\usepackage{url}
\usepackage{verbatim}
\usepackage{graphicx}
\usepackage{cite}
\usepackage{ulem}
\usepackage{threeparttable}
\usepackage[pagebackref,breaklinks,colorlinks,citecolor=blue]{hyperref}
\usepackage[capitalize]{cleveref}
\usepackage{tabularx}
\usepackage{array}
\newcommand{\PreserveBackslash}[1]{\let\temp=\\#1\let\\=\temp}
\newcolumntype{C}[1]{>{\PreserveBackslash\centering}p{#1}}
\newcolumntype{R}[1]{>{\PreserveBackslash\raggedleft}p{#1}}
\newcolumntype{L}[1]{>{\PreserveBackslash\raggedright}p{#1}}

\begin{document}

\title{MWFormer: Multi-Weather Image Restoration Using Degradation-Aware Transformers}

\author{Ruoxi Zhu, Zhengzhong Tu$^{\dag}$, Jiaming Liu, Alan C. Bovik,~\IEEEmembership{Life Fellow,~IEEE} and Yibo Fan$^{\star}$$^{\dag}$,~\IEEEmembership{Member,~IEEE}
\thanks{Ruoxi Zhu, Jiaming Liu, and Yibo Fan are with Fudan University, Shanghai
200433, China. (email: \{rxzhu22@m., liujm22@m., fanyibo@\}fudan.edu.cn). They were supported in part by the National Key R\&D Program of China (2023YFB4502802), in part by the National Natural Science Foundation of China (62031009), in part by Fudan-ZTE joint lab, in part by Alibaba Innovative Research (AIR) Program, in part by Alibaba Research Fellow (ARF) Program.}
\thanks{Zhengzhong Tu is now with the Department of Computer Science and Engineering, Texas A\&M University, College Station, TX 77840, USA (email: tzz@tamu.edu).
This work was done prior to his employment by Texas A\&M University, and he was not supported by any grant.}
\thanks{A. C. Bovik is with the Laboratory for Image and Video Engineering (LIVE) at The University of Texas at Austin, Austin TX 78712, USA (email:bovik@ece.utexas.edu).}
\thanks{$^{\star}$Yibo Fan is the corresponding author. $^{\dag}$: equal advising.}
}

\markboth{Journal of \LaTeX\ Class Files,~Vol.~14, No.~8, August~2021}%
{Shell \MakeLowercase{\textit{et al.}}: A Sample Article Using IEEEtran.cls for IEEE Journals}


\maketitle
\begin{abstract}
Restoring images captured under adverse weather conditions is a fundamental task for many computer vision applications.
However, most existing weather restoration approaches are only capable of handling a specific type of degradation, which is often insufficient in real-world scenarios, such as rainy-snowy or rainy-hazy weather.
Towards being able to address these situations, we propose a multi-weather Transformer, or MWFormer for short, which is a holistic vision Transformer that aims to solve multiple weather-induced degradations using a single, unified architecture.
MWFormer uses hyper-networks and feature-wise linear modulation blocks to restore images degraded by various weather types using the \textit{same} set of learned parameters.
We first employ contrastive learning to train an auxiliary network that extracts content-independent, distortion-aware feature embeddings that efficiently represent predicted weather types, of which more than one may occur.
Guided by these weather-informed predictions, the image restoration Transformer adaptively modulates its parameters to conduct both local and global feature processing, in response to multiple possible weather.
Moreover, MWFormer allows for a novel way of tuning, during application, to either a single type of weather restoration or to hybrid weather restoration without any retraining, offering greater controllability than existing methods.
Our experimental results on multi-weather restoration benchmarks show that MWFormer achieves significant performance improvements compared to existing state-of-the-art methods, without requiring much computational cost. Moreover, we demonstrate that our methodology of using hyper-networks can be integrated into various network architectures to further boost their performance.
The code is available at: \url{https://github.com/taco-group/MWFormer}
\end{abstract}

\begin{IEEEkeywords}
image restoration, adverse weather, multi-task learning, low-level vision, transformer
\end{IEEEkeywords}

\section{Introduction}
\IEEEPARstart{I}{mages} captured in the real world are often of defective quality due to adverse capture or environmental conditions. 
For example, CMOS-based cameras typical in mobile devices often struggle to produce high-quality pictures in low light.
The photos produced under such conditions can be noisy, blurry, and under-exposed.
Other common occurrences of degradation are caused by possibly multiple coincident weather conditions, such as rain, fog, and snow, that affect human-perceived image quality.
When the images are fed to automated vision systems,  these distortions can severely hamper the performances of computer vision algorithms, which are often trained on datasets of pictures taken under normal weather conditions.
Failing to account for and ameliorate the effects of these and other natural phenomena can often lead to catastrophic outcomes in vision-dependent applications like autonomous driving, robotics, security, and surveillance, etc.

\begin{figure}[!t]
    \centering
    \includegraphics[width=0.95\linewidth]{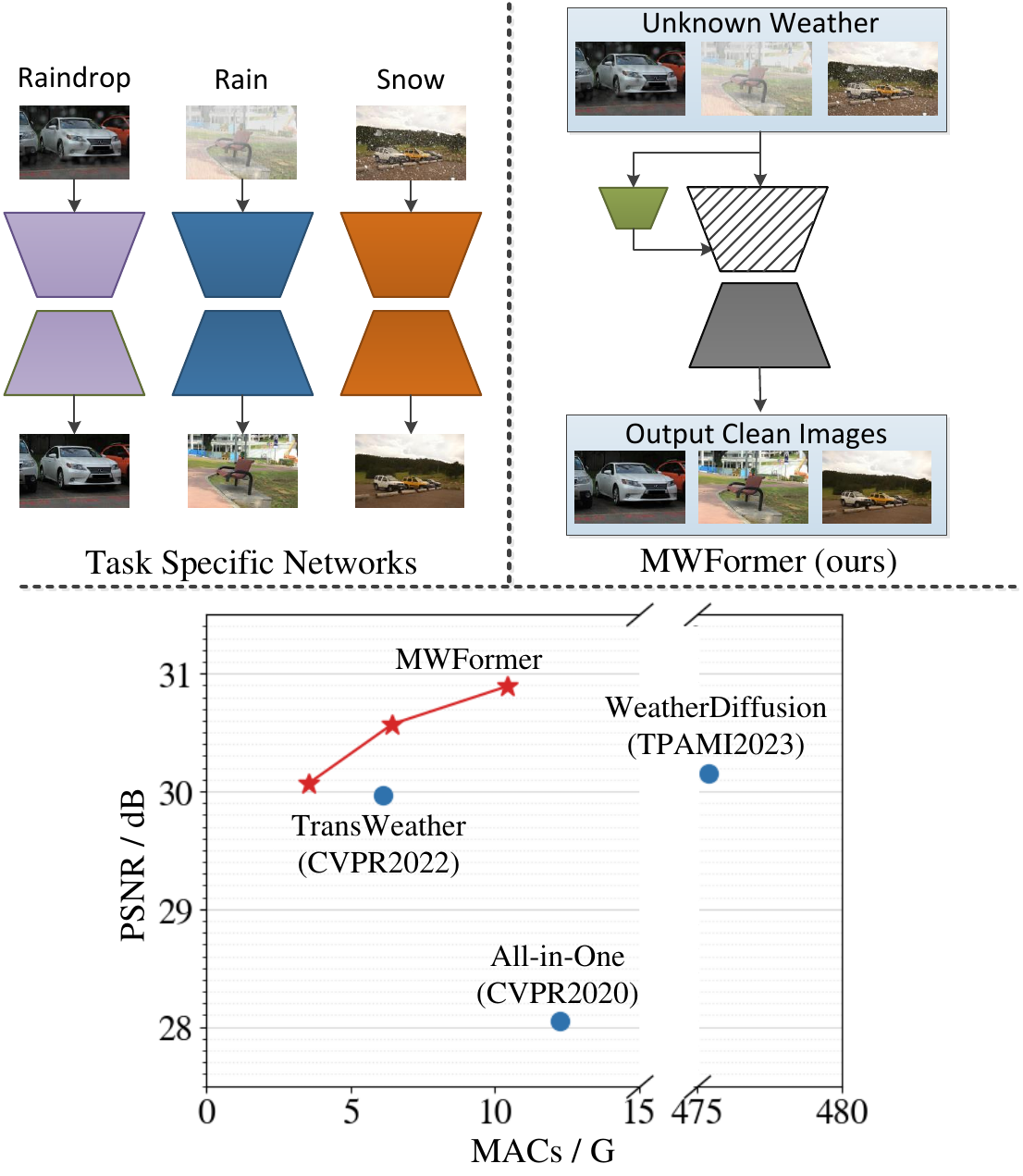}
    \vspace{-0.2cm}
    \caption{Top row: Comparison of the MWFormer architecture with those of existing task-specific networks. Bottom: Restoration performance and computational cost of three versions of MWFormer having different numbers of channels against three competitive multi-weather restoration models. 
    MWFormer achieves generally better performance with \textbf{100$\times$} less computation than the SOTA model WeatherDiffusion~\cite{Diffusion}.
    }
    \label{model_compare}
\end{figure}

Developing image processing algorithms that are able to analyze and subsequently restore weather-degraded pictures is an active research topic ~\cite{TransWeather,li2019single, wu2021contrastive}.
In recent years, deep learning-based restoration methods have been widely utilized to conduct weather-related image restoration tasks, such as deraining \cite{RESCAN, li2019single}, snow-removal \cite{desnow, liu2018desnownet, li2021online}, and dehazing \cite{dehaze,li2018benchmarking,wu2021contrastive}.
Although these methods delivered promising results, each is designed to handle only a single type of adverse weather condition.
Whereas in many real-world scenarios, the weather conditions are generally unknown to the restoration algorithm. 
Moreover, there are often multiple commingled conditions, which result in multiply-distorted pictures that the above-mentioned methods are unable to adequately improve.


Recently, several unified solutions have been proposed to restore images impaired by multiple coincident weather-induced degradations \cite{blindSR, AllInOne, AllInOne2, Diffusion}. 
For example, the authors of \cite{AllInOne, TransWeather, Diffusion} train single networks on combined datasets, each representative of a single weather condition, with the expectation that the models would learn to adaptively process each weather degradation.
However, these methods often deliver unsatisfying and unbalanced generalization performances across different weather types, and are unable to handle artifacts from co-occurring weather conditions.
An important reason for this is that multiple coinciding distortions mutually interact, creating new and highly diverse distortions.

\begin{figure*}[!ht]
    \centering
    \includegraphics[width=1.0\linewidth]{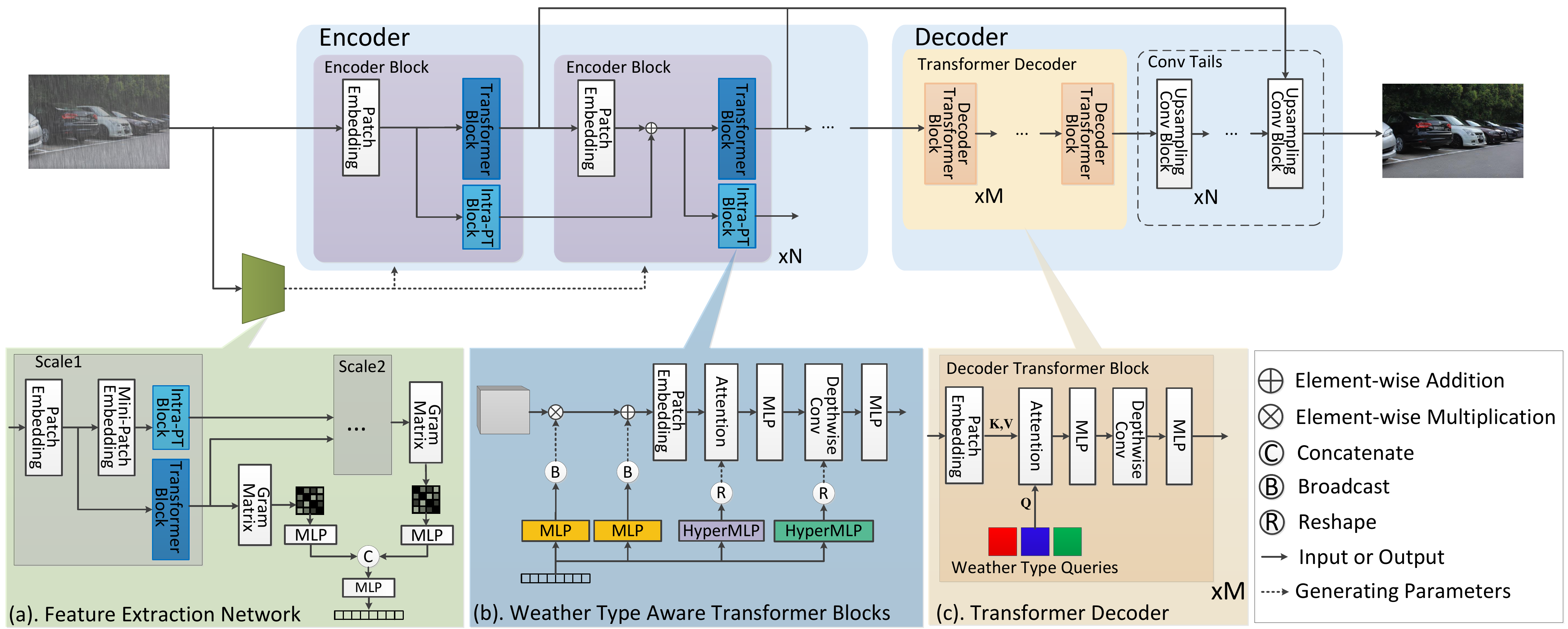}
    \caption{The architecture of MWFormer. The main image processing network consists of a Transformer encoder, a Transformer decoder, and convolution tails.
    (a) A feature extraction network learns to generate some of the parameters of the Transformer blocks and intra-patch Transformer blocks in the main network, thereby partially controlling the production of intermediate feature maps. (b) The Transformer block in the encoder of the main network, which is guided by the feature vector. (c) Transformer decoder of the main network, whose queries are learnable parameters.}
    \vspace{-0.2cm}
    \label{net_arch}
\end{figure*}

Towards making further progress on this important problem, we propose an efficient, degradation-aware \underline{\textbf{M}}ulti-\underline{\textbf{W}}eather Trans\underline{\textbf{Former}} which we call \textbf{MWFormer}, that uses the architecture shown in Fig.~\ref{net_arch}.
MWFormer is designed to provide a strong restoration backbone for conducting image restoration tasks in the presence of unknown adverse weather conditions.
MWFormer is able to account for different weather-induced degradation types using a small auxiliary hyper-network that extracts degradation-informed features from an input image.
These features guide the generation of the parameters of the image restoration backbone, allowing it to adaptively process the picture conditioned on the predicted weather degradation.
We also show that the new hypernet-based multi-weather feature extractor enables a novel way of test-time tuning to either handle a fixed weather condition with less computation, or to handle combined, hybrid weather-induced degradations, without any retraining. This offers greater flexibility and controllability than existing multi-task methods. Notably, the proposed model is the first one capable of handling hybrid-weather degradations that were unseen during training.
Some extended applications of the hyper-network have also been developed, such as identifying the adverse weather type, and guiding the pre-trained weather-specific image restoration models, which shows its versatility. 
Experimental results on benchmark datasets show that MWFormer is able to significantly outperform previous state-of-the-art (SOTA) models, both quantitatively and qualitatively, on a multi-weather restoration benchmark.
Our methodology can also be integrated into various other network architectures to boost their performance in multi-weather restoration.
To sum up, our contributions are summarized as follows:

\begin{itemize}






\item We introduce a novel Transformer-based architecture called MWFormer for multi-weather restoration, which can restore pictures distorted by multiple adverse weather degradations using a single, unified model.

\item A hyper-network is employed to extract content-independent weather-aware features that are used to dynamically modify the parameters of the restoration backbone, allowing for degradation-dependent restoration and other related applications.

\item The feature vector produced by the hyper-network is leveraged to guide the restoration backbone's behavior across all dimensions and scales (i.e., locally spatial, globally spatial, and channel-wise modulations).

\item Two variants of MWFormer are created---one for lower computational cost, and the other for addressing hybrid adverse weather degradations unseen during training.

\item Comprehensive experiments and ablation studies demonstrate the efficacy of the proposed blocks and the superiority of MWFormer in terms of visual and quantitative metrics. We also develop and analyze multi-weather restoration models in the context of downstream tasks.

\end{itemize}

\section{Related Work}
\label{sec:related_work}

\textbf{Image Restoration.} 
Image restoration is a long-standing computer vision problem that aims to reconstruct a high-quality image from a degraded input.
%
Recently, there has been a trend of employing end-to-end training of large neural networks on large-scale paired image datasets for a broad range of tasks, such as denoising~\cite{zhang2017beyond, zhang2018ffdnet}, deblurring~\cite{zhang2022deep,kupyn2018deblurgan,kupyn2019deblurgan}, super-resolution~\cite{ledig2017photo,lim2017enhanced}, low-light enhancement~\cite{guo2016lime,jiang2021enlightengan,meng2020gia}, dehazing~\cite{li2018benchmarking,dehaze}, deraining~\cite{DerainTransformer,AllInOne}, etc.
The impressive advancements on these problems have been mainly driven by the development of novel network architectures.
For example, encoder-decoder architectures have been widely adopted for a wide variety of restoration tasks~\cite{zhang2022deep,kupyn2018deblurgan,kupyn2019deblurgan,jiang2021enlightengan,meng2020gia}, largely because of the efficacy of multi-scale feature learning.
Similarly, the spatial and channel self-attention mechanisms have been used to learn spatially focused and sparser features~\cite{li2018recurrent,li2019single}.
%
More recently, multi-stage progressive networks~\cite{MPRNet,chen2021hinet,maxim} have been deployed on more challenging tasks like deblurring and deraining, achieving impressive performances.

\textbf{Image Deraining.}
Rain can significantly degrade the quality of captured pictures.
Extensive research efforts have aimed to mitigate the adverse effects of rain on images. 
Restoring ``rainy" images involves two sub-tasks: eliminating rain streaks and removing raindrops.
For instance, Li et al.~\cite{RESCAN} leveraged a combination of dilated convolutional neural networks and recurrent neural networks to effectively expunge rain streaks from pictures.
Yasarla et al.~\cite{Syn2Real} utilized a Gaussian Process-based semi-supervised learning framework, demonstrating impressive generalization capabilities on real-world images.
Ba et al.~\cite{GTRain} proposed a novel deraining network trained on a new and comprehensive dataset of real-world rainy images.
Beyond merely addressing rain streaks, there's an increasing emphasis on tackling the challenges posed by raindrops.
Qian et al. \cite{Raindrop} introduced a dataset specifically designed to capture raindrop-related artifacts. They also trained an attentive GAN to effectively remove raindrops. 
Quan et al. \cite{RaindropAndRain} developed a cascaded network designed to simultaneously remove both raindrops and rain streaks.
%
More recently, Xiao et al. \cite{DerainTransformer} developed a Transformer architecture to conduct joint raindrop and rain streak removal, obtaining promising visual results.

\textbf{Image Desnowing.}
Snow is a complex atmospheric phenomenon that plagues the performance of computer vision models, such as the object detectors used in autonomous vehicles.
DesnowNet~\cite{DesnowNet} pioneered the use of deep learning to conduct single-image desnowing, and the authors also built the first ``snowy'' picture dataset, called Snow-100K.
Building on this foundation, Chen et al.~\cite{chen2020jstasr} addressed the \textit{veiling effect}---a phenomenon whereby snowflakes obscure and diminish picture clarity, by proposing a size- and transparency-aware snow removal algorithm.
Recently, Lin et al.~\cite{LMQFormer} designed a lightweight Laplace Mask Query Transformer for snow removal, achieving SOTA performance.

\textbf{Multi-Weather Restoration.} 
The existence of many different weather types in the real world poses a significant challenge to single weather restoration models, leading to growing interest in developing image restoration models that can effectively restore images affected by various complex weather conditions within a single, unified framework. 
%
%
Chen et al.~\cite{two_stage} leveraged a two-stage knowledge learning mechanism to handle three different types of weather with a unified network.
Li et al.~\cite{AllInOne} designed an architecture called All-in-One, equipped with multiple encoders to capture different degradations and a single decoder.
While this approach is promising, its significant computational overhead poses challenges for real-world applicability.
Valanarasu et al.~\cite{TransWeather} unveiled a more efficient Transformer-based architecture called TransWeather by incorporating intra-patch Transformer blocks (intra-PT blocks) and using learnable weather-type queries. The intra-PT blocks share the same architecture as the vanilla Transformer blocks, but take smaller patch embeddings as input, which are sub-patches yielded from the original patch embeddings. These smaller sub-patches facilitate the network to extract finer details which are beneficial for mitigating smaller degradations.
%
%
Ozdenizci et al.~\cite{Diffusion} employed denoising diffusion models to conduct multi-weather image restoration, setting new benchmarks in performance.
Yet, this approach suffers from extremely slow inference time, making it unsuitable for real-time deployments.
Also, the model's design overlooks specific treatments regarding the characteristics of various weather types.
Zhu et al.~\cite{Zhu2023cvpr} proposed a more explainable method to extract the weather-general and weather-dependent features for multi-weather restoration.
Besides, in addition to image restoration models, some researchers \cite{FIFO} have also proposed image segmentation models that can handle different real weather types.

\textbf{Transformers for Image Restoration.}
Building on foundational works \cite{ViT, SwinTransformer}, Transformer architectures have become popular for various computer vision tasks including image restoration, often significantly surpassing the previous CNN-based solutions. 
The Image Processing Transformer (IPT)~\cite{chen2021pre} was the first to employ a pure Transformer architecture for image processing tasks, which was pre-trained on a large number of corrupted image pairs using contrastive learning.
The pre-trained IPT could efficiently adapt to many image processing tasks after fine-tuning, outperforming state-of-the-art methods.
The SwinIR \cite{SwinIR} architecture, based upon the Swin Transformer~\cite{SwinTransformer}, effectively handled low-level vision tasks by leveraging local-attention models.
The Restormer~\cite{Restormer} architecture deployed a novel Transformer variant able to capture long-range pixel interactions while remaining efficient using a transposed attention mechanism.
Furthermore, the Uformer~\cite{Uformer} presented a U-shaped Transformer architecture with locally enhanced windows that has been shown to perform remarkably well across diverse image restoration tasks.


\section{Proposed Method}

Here, we explain the technical details of the proposed MWFormer multi-weather restoration model.
Our primary objective is to learn a single, unified model capable of handling multiple different weather degradations with the same set of learned parameters.
This is similar to the challenge of real-world image denoising, where an algorithm is expected to deal with various noise sources, types, and levels.
%
Non-blind denoising generally outperforms blind denoising, since additional noise information helps a denoising network to better learn to adapt its parameters.
Thus, adding an extra noise estimation module can enhance the performance of the blind denoising network and increase its flexibility.
Drawing inspiration from this, we propose to deem different weather types as analogous to varying noise sources or types.
Features descriptive of weather type can be extracted beforehand, then fed to the main restoration network, which gains degradation adaptivity conditioned on the input weather types.
Consequently, our proposed algorithm may be bifurcated into two phases: weather-feature extraction (by the hyper-network) followed by a weather-type-informed image restoration process.


 \subsection{Overall Architecture}

 An overall schematic diagram of MWFormer is illustrated in Fig.~\ref{net_arch}, showing the two major components: (i) a restoration backbone containing encoder and decoder blocks, which are responsible for recovering a high-quality image from the degraded input; (ii) a feature extraction network that yields weather-aware feature vectors.
 %
 %
 %
We adopt a Transformer-based architecture as the restoration backbone. Besides the vanilla Transformer blocks, our encoder network contains extra intra-PT blocks introduced in Sec. \ref{sec:related_work}.
 The decoder of the backbone is similar to the design in~\cite{TransWeather}, including learnable weather-type queries that cross-attend to the key and value features from the encoders.
 However, this architecture is still incapable of learning to disentangle commingled weather features, arising from coexisting weather conditions, even if it is trained on multiple weather datasets.
 Therefore, we have designed an array of improvements that explicitly supply network flexibility in the multi-weather setting.
 The innovative designs we make are further explained in the following sections.

\subsection{Feature Extraction Network}
\label{ssec:feat_extract}

Weather variations can be viewed as distinct image ``styles'', which are inherently decoupled from the image content.
To illustrate this idea, consider two snapshots of an identical scene, each captured under different weather conditions and manifesting distinct weather-related impairments. Each impaired (or ``weather-styled'') picture should be treated differently by the restoration network, but the two outputs should both faithfully recover the image content.
On the other hand, pictures containing different contents, but suffering from the same weather degradation, should lead to comparable responses from the network.
This is analogous to image style transfer, which emphasizes decoupling image style and content.
The Gram matrix~\cite{gatys2015neural}, which represents correlations within feature maps, is commonly used to define image styles.
Yet, the original form of the Gram matrix fails in the context of multi-weather restoration, as it represents artistic styles rather than weather-relevant features.
To address this, we append trainable projection layers---multi-layer perceptrons (MLPs)---on top of the vanilla Gram matrix, to learn weather-specific ``style''.

The architecture of our feature extraction network is shown in Fig.~\ref{net_arch}(a).
We utilize the first two scales of the Transformer encoders, where a Gram matrix is computed at each scale.
Since Gram matrices are symmetric, only the upper triangular parts of the two matrices are vectorized to save computation. These vectors are further fed to the two projection layers (MLPs), thereby generating two 64-dimensional embeddings.
Finally, the two embeddings are concatenated and projected onto a single feature vector $\bm{v}$, which encodes the weather-degradation information from the input image.

The feature extraction network is intended to cluster images affected by similar weather degradations, hence we utilize contrastive learning~\cite{pmlr-v119-chen20j} to train it, wherein the loss is formulated as:
\begin{equation}
    \mathcal{L}_{con} = \sum_{(a,b)\in\mathcal{P}}\{\mathbb{I}(a,b)[m-d(\bm{v}_a, \bm{v}_b)]_+ + [1-\mathbb{I}(a,b)]d(\bm{v}_a, \bm{v}_b)\},
    \label{feature_loss}
\end{equation}
where $\mathcal{P}$ denotes every possible image pair in a batch, $d(\cdot)$ denotes cosine similarity, $m$ is a positive margin, and $\mathbb{I}(a,b)$ is an indicator that equals 1 when the two images $(a,b)$ contain the same weather impairments and 0 if they are captured under different weather conditions.
The definition of $[\cdot]_+$ operation can be expressed as:
\begin{equation}
    [x]_+ = \begin{cases}
                0, \quad x \leq 0,\\
                x, \quad x > 0.
          \end{cases}
\end{equation}
When calculating the contrastive loss, each possible image pair is sampled from the batch. If the two images belong to two different datasets, the term $d(\bm{v}_a, \bm{v}_b)$ enforces that their feature vectors are pushed away from each other. If the two images belong to the same dataset, the term $[m-d(\bm{v}_a, \bm{v}_b)]_+$ pulls their feature vectors closer in the embedding space. Consequently, the learned feature extraction network is able to cluster the images affected by the same weather degradation.


\subsection{Image Restoration Network}
\label{ssec:image-restoration-network}
The image restoration network contains two sets of learned parameters: fixed parameters that encode the general restoration priors relevant to all the tasks, and weather type-adaptive parameters that are generated by the feature extraction network, as shown in Fig.~\ref{net_arch}(b).
More specifically, the output image $\bm{Y}$ is computed as:
\begin{equation}
    \bm{v}=\mathcal{F}_{feat}(\bm{I}; \tau),
\end{equation}
\begin{equation}
    \bm{Y} = \mathcal{F}_{res}(\bm{I}; \theta_{fix}, \theta_{adap}(\bm{v})),
\end{equation}
where $\mathcal{F}_{feat}$ is the auxiliary feature extraction network (Sec.~\ref{ssec:feat_extract}) with parameters $\tau$, and $\mathcal{F}_{res}$ is the image restoration backbone.
The parameters $\theta_{fix}$ and $\theta_{adap}(\bm{v})$ are the weather-independent and weather-adaptive weights in the encoder stages, respectively.
Since different weather types require varying scales of treatments---for example, deraining mostly requires local contexts, while desnowing demands global understanding to differentiate snowflake and snowpack---we inject weather type adaptivity in multiple pillars: spatial-wise, both locally and globally in the parameter space,  as well as channel-wise feature modulation, to enable better feature learning.
The adaptivity is applied to both the Transformer block and the intra-PT blocks in the encoder stage.
In the Transformer decoder blocks \cite{TransWeather}, the learnable weather-type queries attend to the input features, followed by standard MLP and depth-wise convolution layers, yielding restored output images $\bm{Y}$.


\textbf{Spatially local adaptivity.} Since vanilla Transformer architectures lack inductive biases expressive of local pixel interactions, we add a depthwise convolution layer between the two MLPs in each feed-forward network (FFN) in the Transformer blocks.
Unlike previous models, however, we leverage the predicted weather type features $\bm{v}$ computed by the hyper-network $\mathcal{F}_{feat}$ to generate the parameters of the depthwise convolution layers, so that pictures degraded by different weather types will be processed by different filters adaptively.
The feature vector $\bm{v}$ is fed into a 2-layer projection MLP (named HyperMLP in Fig. \ref{net_arch} since it is intended to generate the parameters of other modules), then reshaped to the 2D depthwise convolution kernels $\bm{w}\in\mathbb{R}^{C\times1\times3\times3}$ (omitting the batch dimension) that are used to convolve the input $\bm{X}_{sl}$:
\begin{equation}  \bm{W}_{DWC}=\text{Reshape}(\text{Proj}(\bm{v})),
\label{eq:dwc}
\end{equation}
\begin{equation} \text{FFN}(\bm{X}_{sl},\bm{v})=\text{MLP}(\sigma(\bm{W}_{DWC}*\bm{X}_{sl})),
\end{equation}
where $\bm{W}_{DWC}$ denotes the weights of the depthwise convolution generated by reshaping the projection of the $\bm{v}$ vector, $\bm{X}_{sl}$ denotes the input of the spatially local operation (i.e., depthwise convolution), $*$ denotes depthwise convolution, and $\sigma$ denotes nonlinear activation.

\textbf{Spatially global adaptivity.} Compared with CNN architectures, Transformers excel in capturing long-range spatial relationships using self-attention layers that scan over all the tokens.
To model adaptive global interactions, we use another hyper-network to compute the critical projecting parameters used in the self-attention operations.
Formally, denoting an input patch embedding of the spatially global operation (i.e., self-attention block) as $\bm{X}_{sg}\in\mathbb{R}^{N\times{C_{in}}}$, three linear projection matrices $\bm{W}_q$, $\bm{W}_k$ and $\bm{W}_v$ are applied to obtain the query, key, and value features $\bm{Q}=\bm{X}_{sg}\bm{W}_q, \bm{K}=\bm{X}_{sg}\bm{W}_k$ and $\bm{V}=\bm{X}_{sg}\bm{W}_v$. The matrix product of $\bm{Q}$ and $\bm{K}^T$ is then calculated, yielding a global attention map to weighted-sum $\bm{V}$.
%
Different weather types may require different attention maps when conducting restoration, thus we employ the weather type feature $\bm{v}$ again, to generate $\bm{W}_q$, $\bm{W}_k$ and $\bm{W}_v$, using a similar projection as our design in Eq.~\eqref{eq:dwc}.
The result is then reshaped to match the dimensions of $\bm{W}\in\mathbb{R}^{d_{in}\times{d_{out}}}$.
Mathematically, 
\begin{equation}  
\bm{W}_i=\text{Reshape}(\text{Proj}(\bm{v})), 
    \quad i=q,k,v,
\end{equation}
\begin{equation}
\bm{Q}=\bm{X}_{sg}\bm{W}_q, \quad \bm{K}=\bm{X}_{sg}\bm{W}_k, 
\quad
\bm{V}=\bm{X}_{sg}\bm{W}_v,
\end{equation}
\begin{equation}
\text{MSA}(\bm{X}_{sg})=softmax(\frac{\bm{Q}\bm{K}^T}{\sqrt{d}})\bm{V}.
\end{equation}

\begin{figure*}
    \centering
    \includegraphics[width=1.0\linewidth]{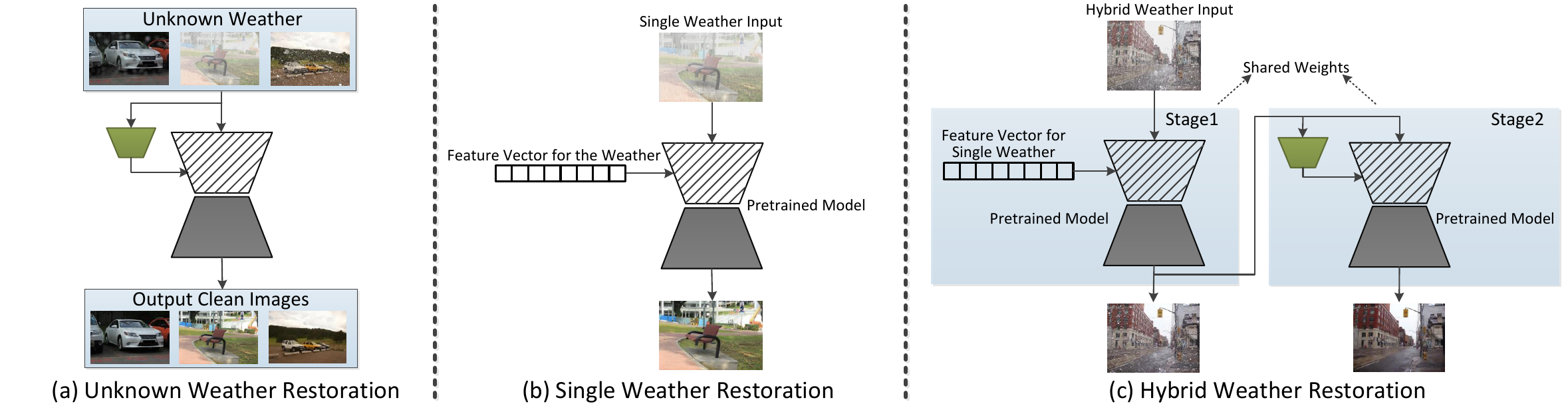}
    \vspace{-0.6cm}
    \caption{Comparison of the default architecture with two test-time variants applied in special cases. To conduct a single weather-type restoration, the feature extraction network is replaced by a fixed feature vector. To conduct hybrid weather restoration, the image processing network is cascaded to remove degradations sequentially, stage by stage.}
    \vspace{-0.4cm}
    \label{variants}
\end{figure*}

\textbf{Channel-wise feature modulation.}
Except for weather-type awareness in the parameter space, we also introduce a dimension of degradation dependency in the intermediate feature space.
We apply a simple affine transformation on the learned intermediate representations, which has been shown to be effective in previous work~\cite{peebles2023scalable, karras2019style, FiLM}.
The feature vector $\bm{v}$ is input to the projection MLPs before each patch embedding layer to generate the weights $\gamma\in\mathbb{R}^C$ and biases $\beta\in\mathbb{R}^C$.
For each channel, the weight and bias are then broadcast to all pixels of the corresponding feature map, which modulates the input features $\bm{X}$ along the channel dimension: 
\begin{equation} \bm{X}'=\gamma\bm{X}+\beta.
\end{equation}
These modulating blocks may be regarded as a form of channel-attention mechanism that re-calibrates the importance of different channels, conditioned on the weather.

\subsection{Simplified Architecture for Fixed Weather Degradation}
%
Besides the aforementioned MWFormer architecture, we also developed a lightweight test-time variant for less computational cost.
Our design for learning representations of weather type using an auxiliary hyper-network, which we use to guide the restoration backbone, also enables a computation-efficient inference scheme when the weather type is already known.
Assuming that the learned weather-representation feature vectors for a given weather type lie near each other in the embedding space, then we can replace the feature extraction network with a fixed feature vector that represents the weather type, which is an approximation of the full-size model.
More specifically, we pre-calculate and store the average feature vector of images affected by each weather type during training, then directly use these features while testing.
This simplified architecture is shown in Fig.~\ref{variants}(b) and formulated as 
\begin{equation}
    \bm{Y} = \mathcal{F}_{res}(\bm{I};\theta_{fix}, \theta_{adap}(\bar{\bm{v}}_i)),
\end{equation}
where $i$ denotes a specific degradation type (e.g., rainstreak, raindrop, snow), and $\bar{\bm{v}}_i$ is the average weather feature vector computed on images affected by the $i$th degradation type during training.
%

\subsection{Multi-stage Architecture for Hybrid Weather Degradations}
\label{hybrid_weather}
We have developed another test-time variant for hybrid adverse weather removal.
Due to a current lack of hybrid weather datasets, previous restoration models, whether trained to handle single or multiple types of weather, are unable to successfully restore pictures captured under multiple simultaneous adverse weather conditions, such as rain + snow.
%
However, MWFormer can be easily modified without re-training to handle previously unseen multiple weather-degraded pictures, hence, it is more generalizable than prior models.
%
%

For example, consider a rain + snow hybrid weather condition.
If a model is only trained on multiple single-weather restoration datasets, then it may be capable of restoring images degraded by any of the weather factors (in this case, rain or snow), but not a combined hybrid weather condition (in this case, rain + snow).
Hence, we develop a two-stage network architecture as a test-time variant of MWFormer to handle such hybrid weather conditions.
In the first stage of inference, the average feature vector for rainy images is used as the guidance of the image restoration backbone to produce an intermediate result that is rain-free but still contains snowflakes.
Then, this intermediate output containing only a single adverse weather is processed using MWFormer again in the second stage to remove snowflakes, yielding a final clean image.
The overall process can be denoted as:

\begin{equation}
    \bm{Z} = \mathcal{F}_{res}(\bm{I};\theta_{fix}, \theta_{adap}(\bar{\bm{v}}_i)),
\end{equation}
\begin{equation}
    \bm{v_z}=\mathcal{F}_{feat}(\bm{Z}; \tau),
\end{equation}
\begin{equation}
    \bm{Y} = \mathcal{F}_{res}(\bm{Z};\theta_{fix},  \theta_{adap}(\bm{v_z})).
\end{equation}

If the image is subjected to more types of adverse weather, then further stages may be cascaded,  wherein each stage restores a specific type of degradation.
Note that the networks in the different stages share the same set of weights, thereby offering flexible test-time augmentation capability without requiring any re-training.

\subsection{Extended Applications}
\label{extended_applications}
The hyper-network that creates weather-informed feature vectors is a key aspect of our approach. Beyond generating parameters and modulating feature maps, these vectors have diverse applications due to the hyper-network's strong perception of weather features. We present two extended applications to demonstrate the versatility of the proposed hyper-network.
\subsubsection{Weather-type identification}
Our hyper-network, trained with a contrastive learning strategy on a multi-weather restoration dataset, contains rich prior information on various weather features. Leveraging this, we developed a weather-type identification method using the hyper-network without re-training.

Taking three adverse weather types as an example. Let $\bm{\bar{v}}_{rd}$, $\bm{\bar{v}}_{r}$ and $\bm{\bar{v}}_{s}$ be the average feature vectors of three weather types (raindrop, rainstreak, and snow) respectively, which is computed during training. To identify the weather type of an image $\bm{I}$ impaired by unknown adverse weather, the image's feature vector $\bm{v}$ is first computed using the feature extraction hyper-network $\mathcal{F}_{feat}$, then the cosine similarities between $\bm{v}$ and the average feature vectors of each weather type are computed:
\begin{equation}
    \bm{v} = \mathcal{F}_{feat}(\bm{I}).
    \label{weather_score1}
\end{equation}
\begin{equation}
    d_i = \frac{\bm{v}\cdot\bm{\bar{v}}_{i}}{\Vert\bm{v}\Vert \Vert\bm{\bar{v}}_{i}\Vert}, \quad i\in\{rd, r, s\}.
    \label{weather_score2}
\end{equation}
Finally, the scores related to each weather type are computed using Softmax function:
\begin{equation}
    s_i = \frac{e^{d_i}}{e^{d_{rd}} + e^{d_r} + e^{d_s}}, \quad i\in\{rd, r, s\}.
    \label{weather_score3}
\end{equation}
The weather score $s_i$ approximately indicates the probability that the image is degraded by the adverse weather type $i$.
If this image is known to be affected by only one of the given adverse weather types, then it can be inferred that the highest-scoring type of weather $i^{*}$ exists in the image:
\begin{equation}
    i^{*} = \arg\max\limits_{i} s_i, \quad i\in\{rd, r, s\}.
\end{equation}
\vspace{-0.5cm}

\subsubsection{Guiding pre-trained weather-specific models}
Most existing adverse weather restoration models are trained for specific weather types, making them effective for known conditions but unable to handle unknown or even hybrid weather scenarios. This limits their real-world applicability. To make full use of these weather-specific experts, we've developed a strategy that utilizes the proposed hyper-network to guide existing pre-trained weather-specific models for restoring images affected by unknown weather conditions.

Suppose we have expert models for many different types of weather. When faced with an image affected by unknown weather conditions, our goal is to select the most suitable expert model, so that the image quality can be improved as much as possible. Without loss of generality, assume that we have three expert models for raindrop removal, rainstreak removal and desnowing respectively. We first compute the weather scores of three weather types using Eq. (\ref{weather_score1}) $\sim$ (\ref{weather_score3}). Then, the highest scoring weather type is considered to be the most typical and most impactful to image quality in this image. 
Therefore, the expert model corresponding to this weather type is selected to process the image. It should be noted that for images affected by hybrid weather, although the degradation may not be completely eliminated, our strategy is able to do as much as possible to improve the image quality with only one pre-trained weather-specific model, whereas other strategies cannot achieve higher image quality with the same or even more computational effort.


\section{Experiments}
In this section, we first detail our experimental settings.
Then, we compare the performance of MWFormer against existing SOTA models both qualitatively and quantitatively.
Furthermore, we also conducted comprehensive ablation studies to study the efficacy of different MWFormer model designs.
Finally, we present some discussions on the effectiveness of the feature vectors in MWFormer and the generalization ability in Sec. \ref{discussions}.

\begin{table*}[ht]
\centering
\begin{threeparttable}
\centering
\setlength{\tabcolsep}{8pt}
\caption{Quantitative comparisons of MWFormer against state-of-the-art multi-weather restoration models on three test datasets. Along each column, the best score is \textbf{boldfaced}, while the other top three are \underline{underlined}.}
\label{comparison}
\begin{tabular}{c|cc|cc|cc|cc|c}
    \toprule
    \multirow{2}{*}{Model} & \multicolumn{2}{c|}{RainDrop~\cite{Raindrop}} & \multicolumn{2}{c|}{Outdoor-Rain~\cite{OutdoorRain}} & \multicolumn{2}{c|}{Snow100K~\cite{Snow100k}}  &  \multicolumn{2}{c|}{Average} & \multirow{2}{*}{MACs (G) $\downarrow$}  \\       & PSNR $\uparrow$    &SSIM $\uparrow$   & PSNR $\uparrow$  & SSIM $\uparrow$   & PSNR $\uparrow$  & SSIM $\uparrow$  & PSNR $\uparrow$  & SSIM $\uparrow$  \\
    \midrule
    AirNet\cite{AllInOne2} & 24.57 & 0.8583 & 18.48 & 0.6719 & 24.41 & 0.8045 & 22.49 & 0.7782 &  301.27 \\
    Chen et al.\cite{two_stage} & \underline{31.83} & \underline{0.9289} & 25.45 & 0.8737 & 28.86 & 0.8886 & 28.71 & 0.8971 & 24.56\\
    All-in-One\cite{AllInOne}       & 31.12 & 0.9268 & 24.71          & 0.8980             & 28.33             & 0.8820 & 28.05 & 0.9023 &  12.26  \\
    TransWeather\cite{TransWeather}     & 30.17             & 0.9157             & 28.83          & 0.9000             & 29.31             & 0.8879 & 29.44 & 0.9012 & 6.13 \\
    WeatherDiffusion\cite{Diffusion} & 30.71             & \textbf{0.9312}    & 29.64        & \textbf{0.9312}    & 30.09 & \underline{0.9041} & 30.15 & \textbf{0.9222} & 475.16 $\times$ 50 \\
    Zhu et al.\cite{Zhu2023cvpr} & 31.31 & \underline{0.93} & 25.31 & 0.90 & 29.71 & 0.89 & 28.78 &  0.91 & 1.36\\
    \midrule
    \textbf{MWFormer-S} (ours)        &31.09 & 0.9224 & 29.07 & 0.9010 & 30.05 & 0.8986 & 30.07 & 0.9073 & 3.57 \\
    \textbf{MWFormer-M} (ours)          &31.56 & 0.9246 & \underline{29.70} & 0.9064 & \underline{30.45} & 0.9029 & \underline{30.57} & 0.9113 & 6.45 \\
    \textbf{MWFormer-L} (ours)          & \underline{31.73}    & 0.9254             & \underline{30.24} & \underline{0.9111} & \underline{30.70}    & \underline{0.9060} & \underline{30.89} & \underline{0.9142} &  10.41  \\
    \midrule
    \textbf{MWFormer-real$^{*}$ (ours)}       & \textbf{31.91} & 0.9268 & \textbf{30.27} & \underline{0.9121} & \textbf{30.92} & \textbf{0.9084} & \textbf{31.03} & \underline{0.9158} & 10.41 \\
    \bottomrule
\end{tabular}
\begin{tablenotes}[para,flushleft]  
        \item $^{*}$MWFormer-real is trained on a larger dataset mentioned in sub-section \ref{training_details}.
\end{tablenotes} 
\end{threeparttable}
\end{table*}

\begin{figure*}[!t]
\centering
\footnotesize
\def\yem{-3pt}
\def\xwidth{1.0}
\def\xxxwidth{0.122\textwidth}
\setlength{\tabcolsep}{1pt}
\begin{tabular}{C{\xxxwidth}C{\xxxwidth}C{\xxxwidth}C{\xxxwidth}C{\xxxwidth}C{\xxxwidth}C{\xxxwidth}C{\xxxwidth}}
\multicolumn{8}{c}{
\includegraphics[width=\xwidth\linewidth]{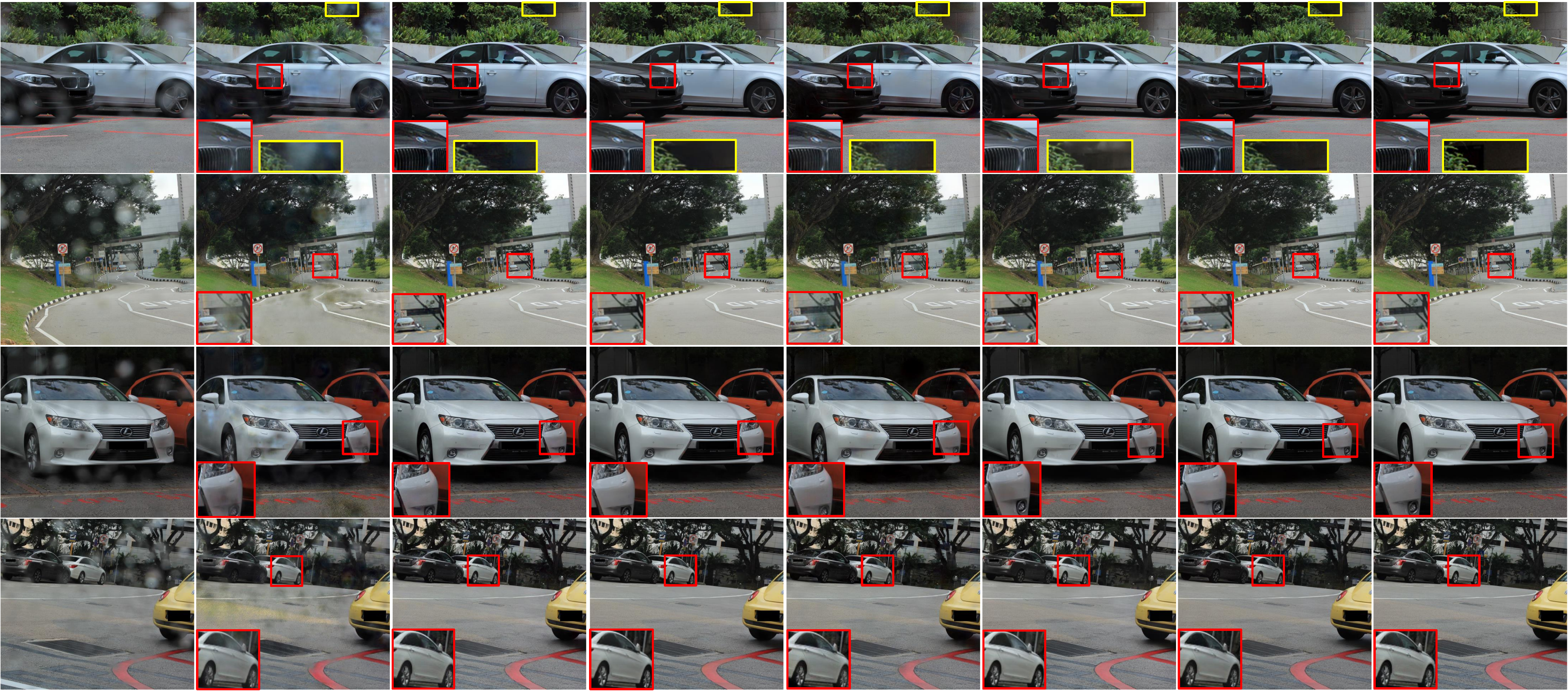}
} \\[\yem]
Input & AirNet~\cite{AllInOne2} & Chen et al.\cite{two_stage} & Zhu et al.\cite{Zhu2023cvpr} & TransWeather~\cite{TransWeather} & WeatherDiffusion~\cite{Diffusion} & MWFormer-L & Ground Truth \\
\end{tabular}
\caption{Qualitative comparisons on the RainDrop \cite{Raindrop} test set. MWFormer effectively removed raindrop artifacts under various scenarios, yielding output images with either fewer shadows or less blur than the other compared models.}
\label{raindrop_compare}
\end{figure*}

\subsection{Training Details}
\label{training_details}
For a fair comparison, we first followed the settings in \cite{AllInOne, TransWeather, Diffusion} to train MWFormer on the standard benchmark for multi-weather restoration, which is a combination of three datasets: RainDrop~\cite{Raindrop}, Outdoor-Rain \cite{OutdoorRain}, and  Snow100K~\cite{Snow100k}. 
%
Similarly, we used the RainDrop test dataset \cite{Raindrop}, the Test1 dataset from Outdoor-Rain \cite{OutdoorRain}, and the Snow100K-L testset~\cite{Snow100k} for testing raindrop removal, draining with dehazing, and desnowing, respectively.

We first pre-trained the feature extraction network in MWFormer over 10k iterations using Eq.~\eqref{feature_loss} as the loss function, with batch size 8 and learning rate $2e^{-4}$. 
Then, we trained the image restoration network over 200k iterations using a weighted combination of the smooth L1 loss and perceptual loss~\cite{perceptual_loss}. In our implementation, the difference between the feature maps extracted by a pretrained VGG16 (from the 3rd, 8th, and 15th layers) of the predicted image and that of the ground truth image was summed up as the perceptual loss.  The total loss function is given as:
\begin{equation}
\mathcal{L}_{all}=\mathcal{L}_1+\lambda\mathcal{L}_{perc},
\end{equation}
where $\lambda$ was fixed at 0.04.
To avoid overfitting to a specific dataset, we sampled approximately the same number of training examples from each dataset, respectively.
Finally, the feature extraction network and the image restoration network were jointly fine-tuned over another 190k iterations using a reduced learning rate.

We instantiated three versions of MWFormer (Small, Medium, and Large), referred to as MWFormer-S, -M, and -L, of our proposed model by changing the number of base channels.
In MWFormer-L, the number of channels for each encoder scale was 64, 128, 320, and 512, respectively, whereas the number of channels was reduced by the factors of 0.75 and 0.5 to create MWFormer-M and MWFormer-S, respectively. 

Besides, it should be noted that some images in this widely adopted benchmark have different distributions from real-world scenes, which is likely to limit the model's real-world performance. For example, this dataset did not represent the veiling effect in multi-weather restoration~\cite{domain_translation}. To further improve MWFormer's applicability to real-world images, we re-trained MWFormer on a larger dataset, which is denoted MWFormer-real. Specifically, besides the previous benchmark dataset, we included another two datasets in the training set: the training set of WeatherStream~\cite{WeatherStream} that contains real-world frames containing rain-fog degradations, and the training set of the CSD dataset~\cite{CSD} that includes images impaired by snowflakes and the veiling effects. We also re-trained TransWeather~\cite{TransWeather} on this larger dataset for a fair comparison.

\begin{figure*}[ht]
    \centering
    \footnotesize
\def\yem{-3pt}
\def\xwidth{1.0}
\def\xxxwidth{0.122\textwidth}
\setlength{\tabcolsep}{1pt}
\begin{tabular}{C{\xxxwidth}C{\xxxwidth}C{\xxxwidth}C{\xxxwidth}C{\xxxwidth}C{\xxxwidth}C{\xxxwidth}C{\xxxwidth}}
\multicolumn{8}{c}{
\includegraphics[width=\xwidth\linewidth]{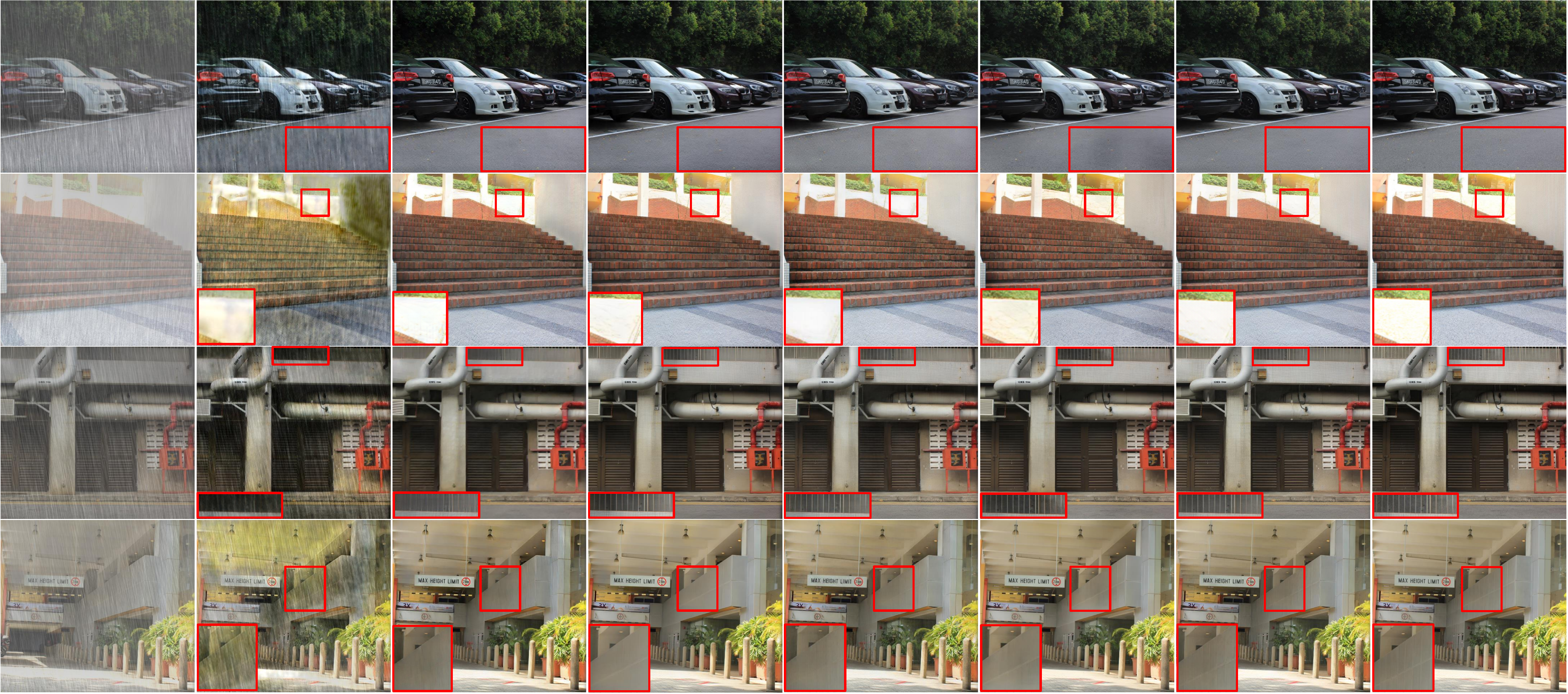}
} \\[\yem]
Input & AirNet~\cite{AllInOne2} & Chen et al.\cite{two_stage} & Zhu et al.\cite{Zhu2023cvpr} & TransWeather~\cite{TransWeather} & WeatherDiffusion~\cite{Diffusion} & MWFormer-L & Ground Truth \\
\end{tabular}
    \caption{Visual comparisons on the Test1 \cite{OutdoorRain} (rain+fog) set. MWFormer performed the best on both detail restoration and luminance retention. AirNet failed to remove most of the degradations, TransWeather recovered fewer details, and the WeatherDiffusion introduced color distortions.}
    \label{rain_compare}
\end{figure*}

\begin{figure*}[ht]
    \centering
    \footnotesize
\def\yem{-3pt}
\def\xwidth{1.0}
\def\xxxwidth{0.122\textwidth}
\setlength{\tabcolsep}{1pt}
\begin{tabular}{C{\xxxwidth}C{\xxxwidth}C{\xxxwidth}C{\xxxwidth}C{\xxxwidth}C{\xxxwidth}C{\xxxwidth}C{\xxxwidth}}
\multicolumn{8}{c}{
\includegraphics[width=\xwidth\linewidth]{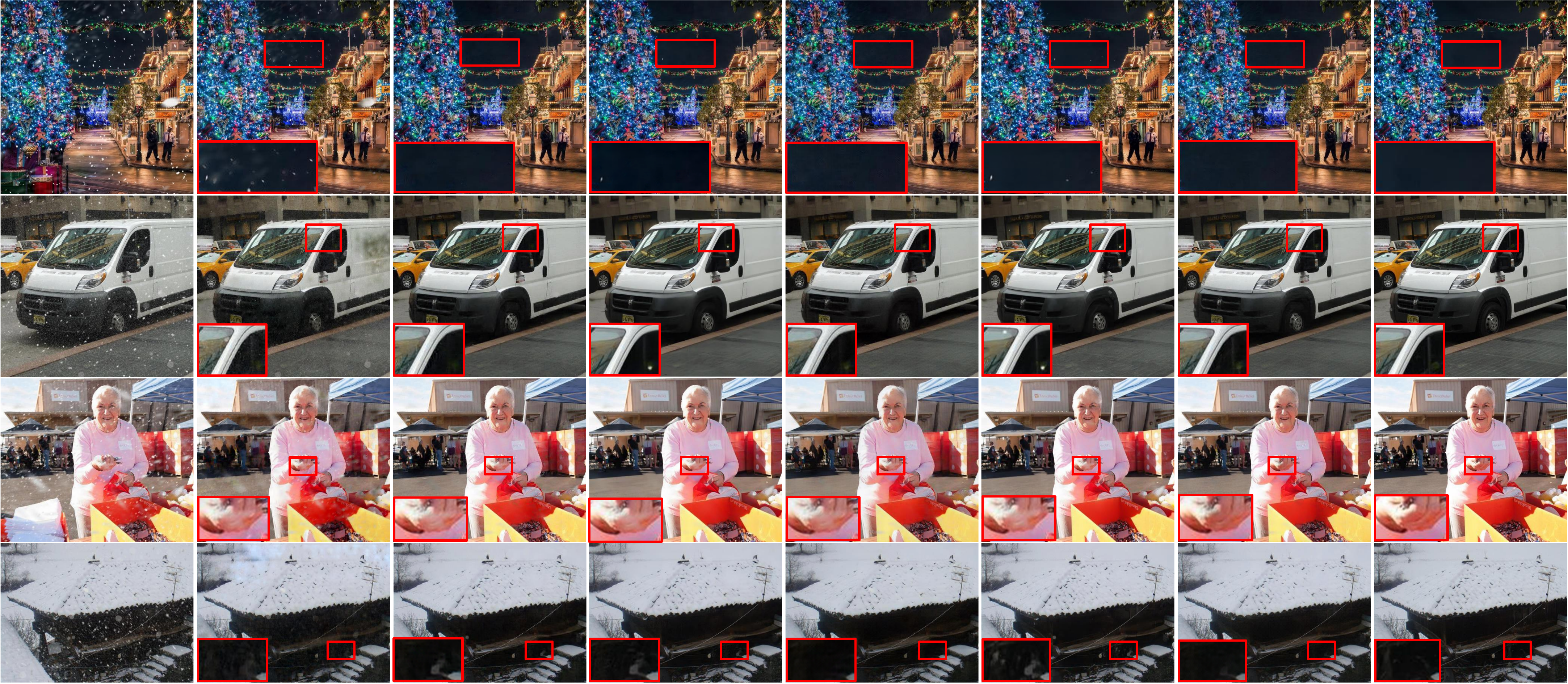}
} \\[\yem]
Input & AirNet~\cite{AllInOne2} & Chen et al.\cite{two_stage} & Zhu et al.\cite{Zhu2023cvpr} & TransWeather~\cite{TransWeather} & WeatherDiffusion~\cite{Diffusion} & MWFormer-L & Ground Truth\\
\end{tabular}
    \caption{Visual comparisons on the Snow100K-L \cite{Snow100k} testset. MWFormer efficaciously removed snowflakes, delivering cleaner pictures than the other models. }
    \label{snow_compare}
\end{figure*}

\subsection{Quantitative Comparisons}
\label{ssec:quantitative_comparison}
We used five state-of-the-art multi-weather restoration models as comparisons: All-in-One\cite{AllInOne}, Chen et al.\cite{two_stage}, TransWeather\cite{TransWeather}, WeatherDiffusion\cite{Diffusion} and Zhu et al.\cite{Zhu2023cvpr}. Another all-in-one image restoration model named AirNet\cite{AllInOne2} was also re-trained on the benchmark dataset for comparison.
Table~\ref{comparison} reports the performances using PSNR and SSIM~\cite{wang2004image} as performance metrics. 
The computational costs of each model, evaluated by the number of multiply-accumulate operations (MACs), are also listed.
As may be seen from the table, MWFormer-real performed best on all three datasets among all the compared methods in terms of PSNR, which is usually regarded as the most reliable measure of fidelity. MWFormer-L also performed better than any model trained using the benchmark dataset regarding average PSNR.
Though Chen et al.\cite{two_stage} achieved better results on the Raindrop testset, their model performed poorly under the other two weather conditions, and the imbalance performance is not preferable in practice.
In terms of the more perceptual-oriented metric SSIM, the diffusion-based WeatherDiffusion model, on average, achieved the best scores, but MWFormer yielded comparable results, performing among the top three.

Although WeatherDiffusion~\cite{Diffusion} performed well in terms of SSIM on some datasets, it requires 2000$\times$ more computation than our largest model MWFormer-L, and requires 5000$\times$ more computation than our smallest model MWFormer-S, if the iterative sampling diffusion process is considered.  Overall, our MWFormer appears to deliver the best trade-off between image quality and computational cost.

Besides, although WeatherDiffusion delivered the best SSIM results on the RainDrop and Outdoor-Rain sets, the diffusion model is occasionally prone to hallucinative artifacts.
One of its failure cases is shown in the third row of Fig. \ref{real_compare}, which exhibits unacceptable artifacts and stains that significantly alter the image contents.   
Since these restoration models are often employed as preprocessing modules for many downstream recognition tasks, such as object detection and semantic segmentation for autonomous vehicles, hallucinations of image content obtained from diffusion-based models could lead to hazardous outcomes in real-world scenarios.

Moreover, the comparison results of TransWeather-real and MWFormer-real are illustrated in Table. \ref{additional_compare}, indicating that MWFormer still surpasses the existing leading models, such as TransWeather, if they are both trained on a larger dataset. Also, by including more images closer to the real scene, the quantity metrics on all of the test sets are boosted.
\begin{table*}[t]
    \centering
    \setlength{\tabcolsep}{8pt}
    \caption{Comparisons of performances of MWFormer-real and TransWeather-real.}
    \begin{tabular}{c|c|c|c|c|c}
    \toprule
    Model & RainDrop~\cite{Raindrop} & Outdoor-Rain~\cite{OutdoorRain} & Snow100K~\cite{Snow100k} & WeatherStream~\cite{WeatherStream} & CSD~\cite{CSD} \\
    \midrule
    TransWeather-real~\cite{TransWeather} & 30.99 / 0.9207 & 28.98 / 0.9002 & 30.00 / 0.8996 & 24.29 / 0.7468 & 32.99 / 0.9580\\
    \textbf{MWFormer-real} (ours) & 31.91 / 0.9268 & 30.27 / 0.9121 & 30.92 / 0.9084 & 24.44 / 0.7488 & 34.60 / 0.9690 \\
    \bottomrule
    \end{tabular}
    \label{additional_compare}
\end{table*}

\begin{figure}[t]
    \centering
    \scriptsize
\def\yem{-3pt}
\def\xwidth{0.99}
\def\xxxwidth{0.118\textwidth}
\setlength{\tabcolsep}{1pt}
\begin{tabular}{C{\xxxwidth}C{\xxxwidth}C{\xxxwidth}C{\xxxwidth}}
\multicolumn{4}{c}{
\includegraphics[width=\xwidth\linewidth]{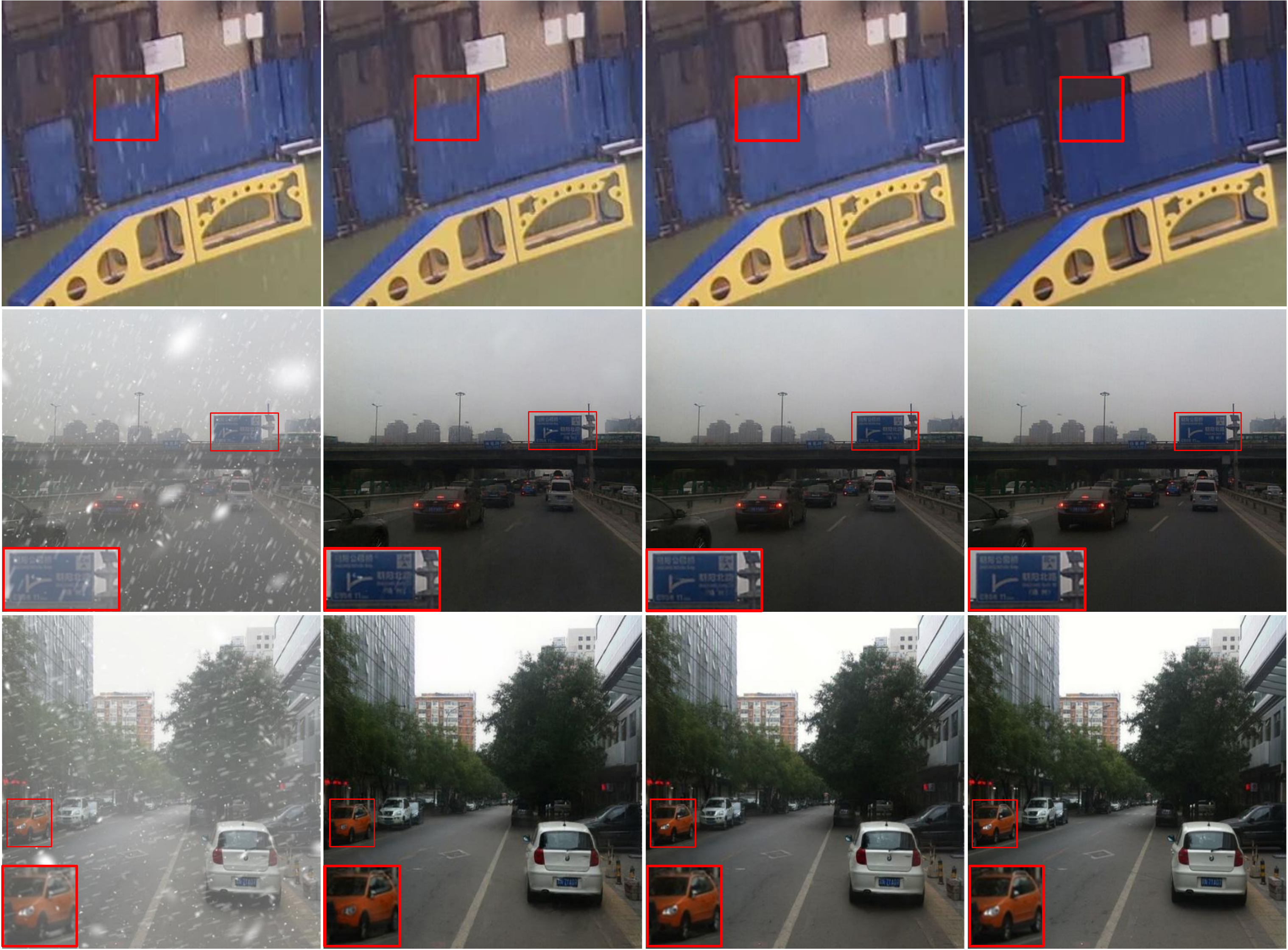}
} \\[\yem]
Input & TransWeather-real & MWFormer-real & Ground Truth \\
\end{tabular}
    \caption{Visual comparisons of TransWeather-real~\cite{TransWeather} and MWFormer-real on two additional testsets (WeatherStream testset ~\cite{WeatherStream} and CSD testset~\cite{CSD}) that are more consistent with real-world scenes. }
    \label{ws_csd}
\end{figure}

\begin{figure*}[ht]
    \centering
    \footnotesize
\def\yem{-3pt}
\def\xwidth{0.99}
\def\xxxwidth{0.16\textwidth}
\setlength{\tabcolsep}{1pt}
\begin{tabular}{C{\xxxwidth}C{\xxxwidth}C{\xxxwidth}C{\xxxwidth}C{\xxxwidth}C{\xxxwidth}}
\multicolumn{6}{c}{
\includegraphics[width=\xwidth\linewidth]{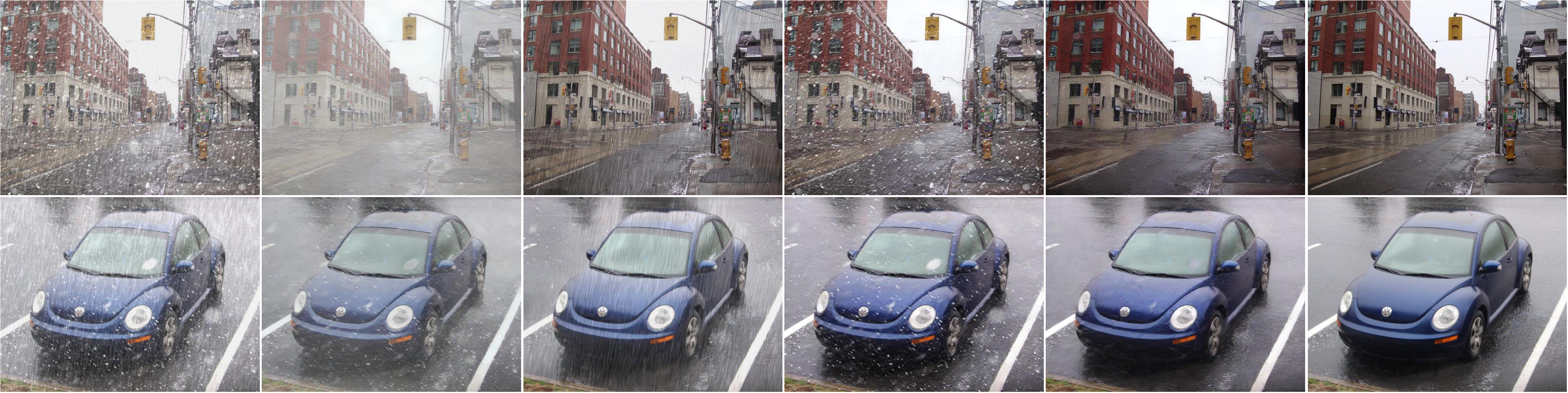}
} \\[\yem]
Input & TransWeather~\cite{TransWeather} & WeatherDiffusion~\cite{Diffusion} & MWFormer (stage1) & MWFormer (2 stages) & Ground Truth \\
\end{tabular}
    \caption{Visual comparisons on hybrid-weather degradations. While most of the compared models failed to handle the complex degradations, the 2-stage MWFormer model, which sequentially removes rain streaks and snowflakes in each stage, was able to deliver more visually appealing outcomes. }
    \label{hybrid_comp}
\end{figure*}

\begin{figure*}[ht]
    \centering    \footnotesize
\def\yem{-3pt}
\def\xwidth{0.99}
\def\xxxwidth{0.140\textwidth}
\setlength{\tabcolsep}{1pt}
\begin{tabular}{C{\xxxwidth}C{\xxxwidth}C{\xxxwidth}C{\xxxwidth}C{\xxxwidth}C{\xxxwidth}C{\xxxwidth}}
\multicolumn{7}{c}{
\includegraphics[width=\xwidth\linewidth]{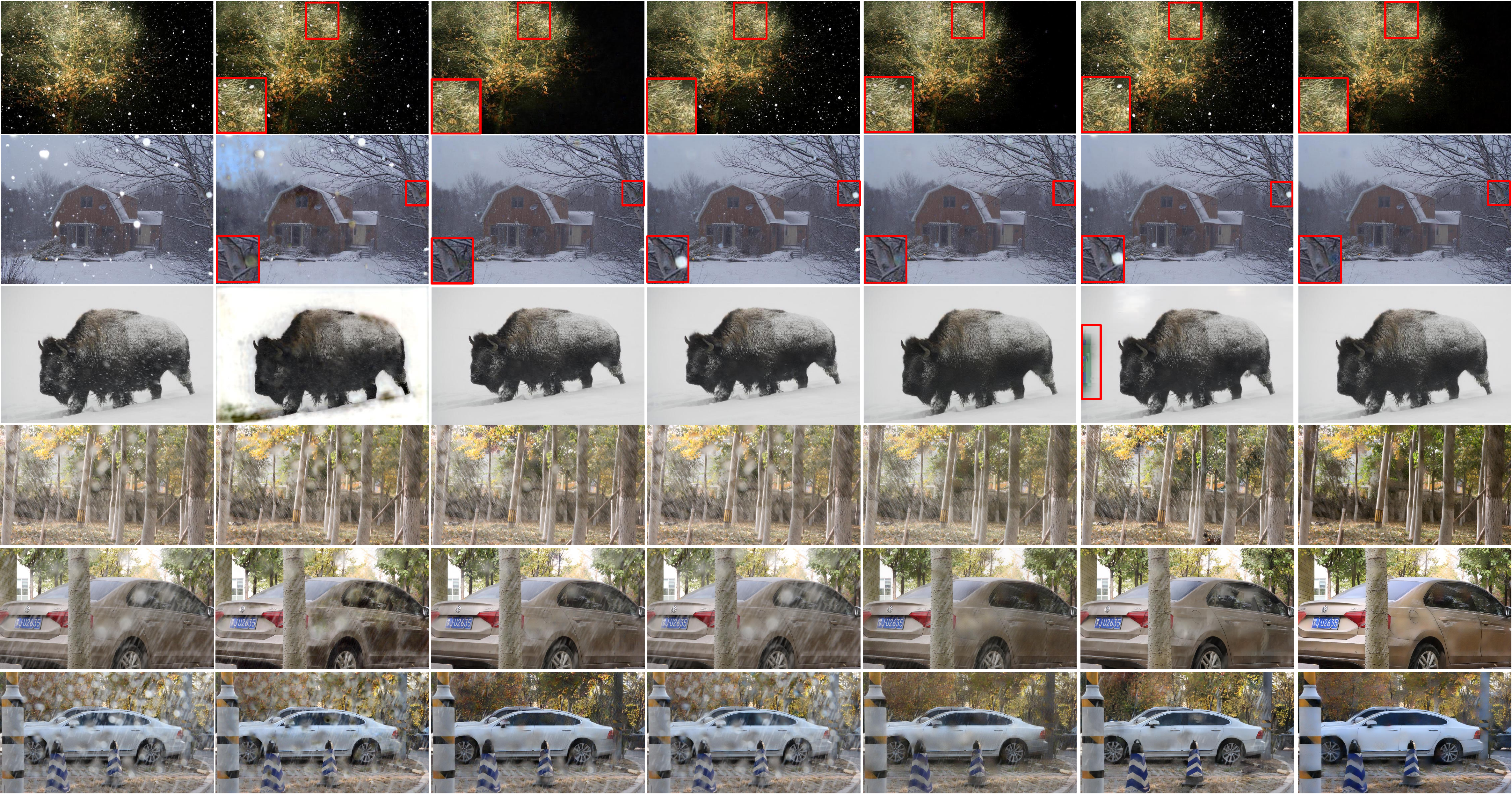}
} \\[\yem]
Input & AirNet~\cite{AllInOne2} & Chen et al.\cite{two_stage} & Zhu et al.\cite{Zhu2023cvpr} & TransWeather~\cite{TransWeather} & WeatherDiffusion~\cite{Diffusion} & MWFormer (ours)\\
\end{tabular}
    \caption{Qualitative results on real images (including hybrid-weather-degraded images) from \cite{Snow100k} and \cite{RainDS}. MWFormer was able to remove the snowflakes while preserving the original image structure. However, AirNet \cite{AllInOne2} and WeatherDiffusion \cite{Diffusion} generated undesirable artifacts. Furthermore, MWFormer can capably remove the hybrid-weather degradations that were unseen during training, as shown in the last three rows.}
    \label{real_compare}
\end{figure*}

\begin{figure*}[ht]
    \centering
     \footnotesize
    \def\yem{-3pt}
\def\xwidth{0.99}
\def\xxxwidth{0.195\textwidth}
\setlength{\tabcolsep}{1pt}
\begin{tabular}{C{\xxxwidth}C{\xxxwidth}C{\xxxwidth}C{\xxxwidth}C{\xxxwidth}}
\multicolumn{5}{c}{
\includegraphics[width=\xwidth\linewidth]{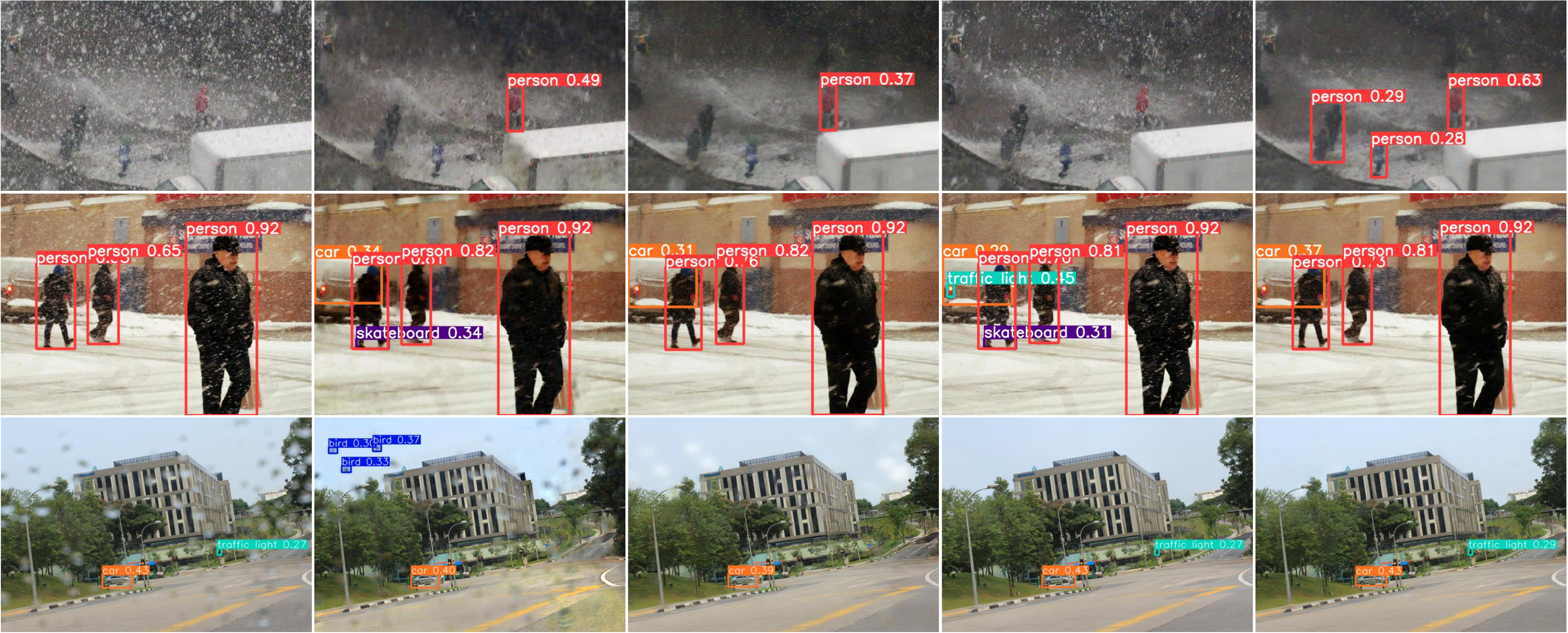}
} \\[\yem]
Input & AirNet~\cite{AllInOne2} & TransWeather~\cite{TransWeather} & WeatherDiffusion~\cite{Diffusion} & MWFormer (ours)\\
\end{tabular}
    \caption{Task-driven comparisons on YOLO-V5 object detection. MWFormer helped deliver better detection performance than other compared methods. Note that AirNet \cite{AllInOne2} and WeatherDiffusion \cite{Diffusion} were implicated in causing false positives in the detection results, likely due to inadequate restoration performance.}
    \vspace{-0.2cm}
    \label{detection_comp}
\end{figure*}

\subsection{Qualitative Comparisons}
\label{ssec:qualitative_comparison}
We also obtained the visual results on each benchmark dataset as shown in~\cref{raindrop_compare,rain_compare,snow_compare}. 
On the \textbf{RainDrop} test dataset, as shown in~\cref{raindrop_compare}, AirNet failed to remove many of the raindrops. Both TransWeather and WeatherDiffusion produced artifacts such as shadows and hallucinations (see the first two rows). MWFormer, however, delivered visually pleasing results without shadows or blur.
On the \textbf{Test1} (rain+fog) dataset shown in Fig. \ref{rain_compare}, MWFormer was able to restore both the luminance and detail information accurately, while the results from Chen et al. and TransWeather suffered from a loss of detail (note the texture in the last two rows), and the results produced by Zhu et al. and WeatherDiffusion included shadows (see the first row). Additionally, WeatherDiffusion sometimes led to color distortion (see the second row).
On the \textbf{Snow100K-L} dataset shown in Fig. \ref{snow_compare}, MWFormer yielded cleaner images, while AirNet, Zhu et al. and WeatherDiffusion tended to interpret some snowflakes as other image details and incorrectly preserved them, thereby reducing the image quality.

We also compared MWFormer-real and TransWeather-real on the two more realistic testsets: WeatherStream~\cite{WeatherStream} and CSD~\cite{CSD} testsets. The visual results are shown in Fig. \ref{ws_csd}. On the WeatherStream dataset, MWFormer-real removes the rainstreaks more thoroughly than TransWeather-real, leading to more visually pleasing results. On the CSD dataset, TransWeather-real sometimes wrongly retains the snowflakes and tends to blur the small but bright objects excessively.

\subsection{Performance on Hybrid Weather Degradations}
More challenging but frequent scenarios are hybrid weather conditions.
Hence, we also studied the performances of the compared models on hybrid-weather-degraded images.
Using the weather synthesizing algorithms in \cite{OutdoorRain}, we simulated images with hybrid degradations of rain + snow using images from Snow100K.
The results of restoring these degraded images are shown in~\cref{hybrid_comp}.
It can be seen that previous models failed to restore these images since obvious snowflakes, rain streaks, or fog remain in their outputs. 
This may be because hybrid-weather-degraded images were not part of their training data; models trained on single weather types cannot be expected to generalize to restore more complex weather degradations.
However, MWFormer, which is imbued with the flexibility to conduct test-time augmentation (\cref{variants}), is able to remove rain and snowflakes in two successive stages, yielding clean, degradation-free images.
We also demonstrate the efficacy of multi-stage application by visualizing the effects of the stage-by-stage degradation removal process in~\cref{hybrid_comp}.

To study the alternative approaches of our proposed multi-stage MWFormer architecture for rain + snow restoration problem, we compared four different strategies: 
First, we applied the simplest single-stage architecture that is intended for single-weather restoration to the rain + snow problem. Second, we applied the single-stage model to each image twice in succession. Third, using a two-stage MWFormer, we conducted desnowing first, using the average feature vector as guidance, followed by deraining. Last, we reversed the order by deraining first and then desnowing, as shown in~\cref{variants}(c).
The performances of these models were tested on our synthetic dataset, which consists of diverse scenes, different rain levels, and different angles of rain streaks.
Quantitative comparisons in Table~\ref{hybrid_weather_abl} indicate that MWFormer performed best when deraining in the first stage, followed by desnowing in the second stage.
This may be because snowflake appearance is significantly affected by rain accumulation; thus, the average feature vector for snow-degraded images may not match these images well.
The intermediate results after deraining resemble the snow-degraded images in the training set, which are easier for the network to process.
These results powerfully demonstrate the efficacy of the MWFormer model in dealing with multi-weather scenarios. 

\begin{table}[t]
    \centering
    \setlength{\tabcolsep}{8pt}
    \caption{Comparisons of performances of three different MWFormer models on hybrid rain + snow images.}
    \begin{tabular}{c|c|c}
    \toprule
    Strategy & PSNR & SSIM \\
    \midrule
    Single Stage & 21.24 & 0.7237 \\
    Two Stages with Default Settings & 22.76 & 0.7665 \\
    Two Stages, Desnow First & 21.98 & 0.7568 \\
    Two Stages, Derain First & 24.80 & 0.7669\\
    \bottomrule
    \end{tabular}
    \label{hybrid_weather_abl}
\end{table}

\subsection{Generalization to Real Weather Degradations}
We also compared MWFormer against other models on the real weather-degraded images from the Snow100K-real set~\cite{DesnowNet} that contains pictures taken under real snowy conditions, and from RainDS-real dataset~\cite{RainDS} that includes real-world images with raindrops and rainstreaks. We used the MWFormer-L to process the images in the Snow100K-real dataset, and use its variant discussed in Sec. \ref{hybrid_weather} to restore the hybrid-weather-degraded images in the RainDS-real dataset.
Note that no ground truth is available for these images, so we must rely on visual comparisons.
As may be seen in~\cref{real_compare}, MWFormer was able to remove most of the snowflakes, yielding visually clean reconstructions as compared to other methods.
As for images impaired by both raindrops and rainstreaks, MWFormer also performed the best, owing to its flexibility for hybrid-weather degradation removal.
Moreover, it should be observed that WeatherDiffusion was exceedingly sensitive to domain shift---its performance varied significantly on different images, and it randomly generated unacceptable artifacts (the third row of~\cref{real_compare}).
MWFormer, on the other hand, produced more visually consistent results in terms of real-weather generalization, which may be attributed to the smaller number of learnable parameters and the design of the weather-type feature learning.
\begin{table}[t]
    \centering
    \small
     \setlength{\tabcolsep}{4pt}
    \caption{NIQE scores of MWFormer and previous SOTA methods on real-world datasets \cite{Snow100k, RainDS}.}
    \vspace{-0.2cm}
    \begin{tabular}{c|c|c|c}
     \toprule
         & TransWeather \cite{TransWeather} & WeatherDiffusion \cite{Diffusion} & MWFormer \\
        \toprule
         NIQE $\downarrow$ &  3.2550 & 3.0162 & \textbf{2.9469}  \\
         \bottomrule
    \end{tabular}
    \vspace{-0.4cm}
  \label{NIQE}
\end{table}
The quantitative comparisons using NIQE \cite{NIQE} are reported in Table \ref{NIQE}, indicating that MWFormer outperforms the previous state-of-the-art models on this most widely adopted no-reference metric.

\subsection{Task-Driven Comparisons}
\label{task_driven_comparisons}
Image restoration results may be consumed either by humans or by machines. It is likely that weather degradation removal is more frequently used in machine vision systems, e.g., as a precursor to object detection for autonomous driving. 
%
We studied this aspect by conducting a study of task-driven image restoration performance in the context of object detection.
Specifically, we evaluated the object detection performance of a pre-trained YOLO-V5~\cite{Jocher_YOLOv5_by_Ultralytics_2020} object detector on images restored by the compared models. 
As shown in \cref{detection_comp}, on real images containing snowflakes, the pictures processed by MWFormer were able to better boost the detection performance of the YOLO-V5 as compared to applying the object detector to the original snow-degraded pictures.
This suggests the potential of using MWFormer as a pre-processing component before object detectors in applications such as Autopilot~\cite{Tesla}.
The other image restoration methods, however, led to fewer detected objects and even misclassified some objects.
It is worth noting that on images affected by raindrops, MWFormer only delivered slightly better detection performance than on the original image, while the other approaches had little effect or even deteriorated the detection performance.
This observation is consistent with the empirical results in~\cite{li2019single}.
Lastly, we observed that AirNet and WeatherDiffusion tended to cause false positive cases (``skateboard'' in the second row and ``bird'' in the bottom row) on some images, which could lead to unexpected and undesirable outcomes in real-world applications.

\begin{figure}[!t]
    \centering
    \footnotesize
    \def\yem{-3pt}
\def\xwidth{0.99}
\def\xxxwidth{0.155\textwidth}
\setlength{\tabcolsep}{1pt}
\begin{tabular}{C{\xxxwidth}C{\xxxwidth}C{\xxxwidth}}
\multicolumn{3}{c}{
\includegraphics[width=\xwidth\linewidth]{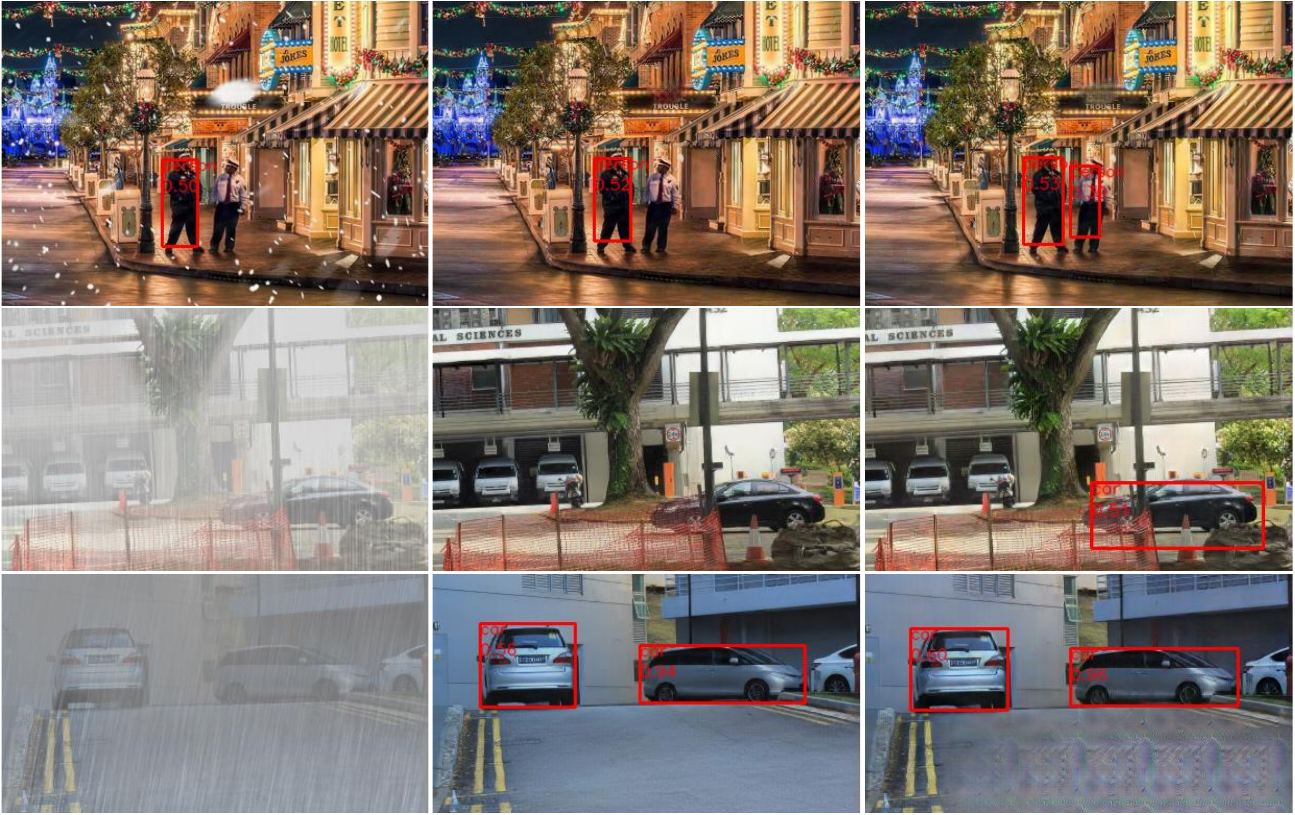}
} \\[\yem]
Input & Visual-Oriented & Detection-Oriented\\
\end{tabular}
    \caption{Comparisons of the effects of visual-oriented and recognition-aware training. It may be observed that the latter strategy yielded less visually appealing outcomes, but led to better detection accuracy.}
    \label{detection_training}
\end{figure}

\textbf{Recognition-aware training.} 
We also studied ways to show that MWFormer can be specifically trained to benefit downstream detection models. 
To do this, we created a recognition-aware version of MWFormer by fine-tuning the base model of MWFormer for a few more steps, replacing the perceptual loss with a recognition loss (including both classification loss and regression loss), calculated using a MobileNetV3-SSDLite object detection network $\mathcal{R}$ with frozen weights.  
To calculate the recognition loss, we assigned the detection results of the clean image $\mathcal{R}(\bm{I}_{clean})$ as ground truth, and calculated its distance to the detection results on each restored image $\mathcal{R}(\mathcal{F}_{res}(\bm{I}; \theta_{fix}, \theta_{adap}(\bm{v})))$. The total loss was: 

\begin{equation}
\mathcal{L}_{all}=\mathcal{L}_1+\lambda(\mathcal{L}_{cls}+\mathcal{L}_{reg}),
\end{equation}
where $\mathcal{L}_{cls}+\mathcal{L}_{reg}$ is the recognition loss consisting of two terms: $\mathcal{L}_{cls}$ is the classification loss implemented as the cross entropy between the predicted logits and the ground truth labels,  and $\mathcal{L}_{reg}$ is the regression loss implemented as smooth L1 loss between the predicted bounding box and the ground truth.

Several visualizations of the restored images with detection results overlaid are shown in~\cref{detection_training}.
The detected object bounding boxes are overlaid, along with their associated detection confidence score.
%
Some interesting observations can be drawn from this experiment. First, including the task-oriented training objective improved the performance of the downstream detection tasks, consistent with the findings in~\cite{9796589}.
Further, optimizing for human quality perception and machine tasks led to different visual effects in the output images, indicating that deep neural network-based detectors learn different representations compared to human visual systems.
Exploring more task-oriented image restoration techniques is beyond the scope of this paper, and hence, we leave it to future work. 

\begin{table*}[ht]
\centering
\setlength{\tabcolsep}{8pt}
\caption{Ablation studies of components of the MWFormer-L architecture.}
\label{ablation}
\begin{tabular}
{cccc|cc|cc|cc|cc}
     \toprule
     \multirow{2}{*}{Local} &\multirow{2}{*}{Global}  &\multirow{2}{*}{Channel} & \multirow{2}{*}{Fine-Tune}  & \multicolumn{2}{c|}{RainDrop \cite{Raindrop}} & \multicolumn{2}{c|}{Outdoor-Rain \cite{OutdoorRain}} & \multicolumn{2}{c|}{Snow100K \cite{Snow100k}} &  \multicolumn{2}{c}{Average}  \\ & & &  & PSNR $\uparrow$    &SSIM $\uparrow$   & PSNR $\uparrow$  & SSIM $\uparrow$   & PSNR $\uparrow$  & SSIM $\uparrow$ & PSNR $\uparrow$  & SSIM $\uparrow$ \\
     \midrule
     \          & \          & \          & \          & 30.72 & 0.9173 & 29.23 & 0.9007 & 30.07 & 0.8992 & 30.01 & 0.9057\\
     \checkmark & \          & \          & \          & 31.11 & 0.9222 & 29.70 & 0.9028 & 30.14 & 0.8998 & 30.32 & 0.9083\\
     \checkmark & \checkmark & \          & \          & 31.19 & 0.9236 & 29.80 & 0.9055 & 30.18 & 0.9010 & 30.39 & 0.9100\\
     \checkmark & \checkmark & \checkmark & \          & 31.36 & 0.9235 & 29.89 & 0.9073 & 30.50 & 0.9041 & 30.58 & 0.9116 \\
     \checkmark & \checkmark & \checkmark & \checkmark & 31.73 & 0.9254 & 30.24 & 0.9111 & 30.70 & 0.9060 & 30.89 & 0.9142 \\
     \bottomrule
\end{tabular}
\end{table*}

\subsection{Ablation Studies}
To further understand and validate the efficacy of MWFormer, we conducted several comprehensive ablation studies. 
%
We used the MWFormer-L as the base model and ablated the various components trained using the same set of hyperparameters.
%
We first trained a baseline MWFormer-L model (without the feature learning network) and gradually added 1) spatially local adaptivity, 2) spatially global adaptivity, 3) channel-wise feature modulation, and 4) joint fine-tuning, as explained in~\cref{ssec:image-restoration-network}.
As may be seen in Table. \ref{ablation}, each axis of weight adaptivity contributed to a notable performance gain on all the datasets, with Local adaptivity delivering the greatest gain on Raindrop and Outdoor-Rain datasets, while Channel adaptivity supplied the most benefit on the Snow100K.
The final stage of joint fine-tuning can further boost the overall performance by aligning the separately trained feature extraction network with the image restoration backbone. 

\begin{figure}[t]
    \centering
    \includegraphics[width=1.0\linewidth]{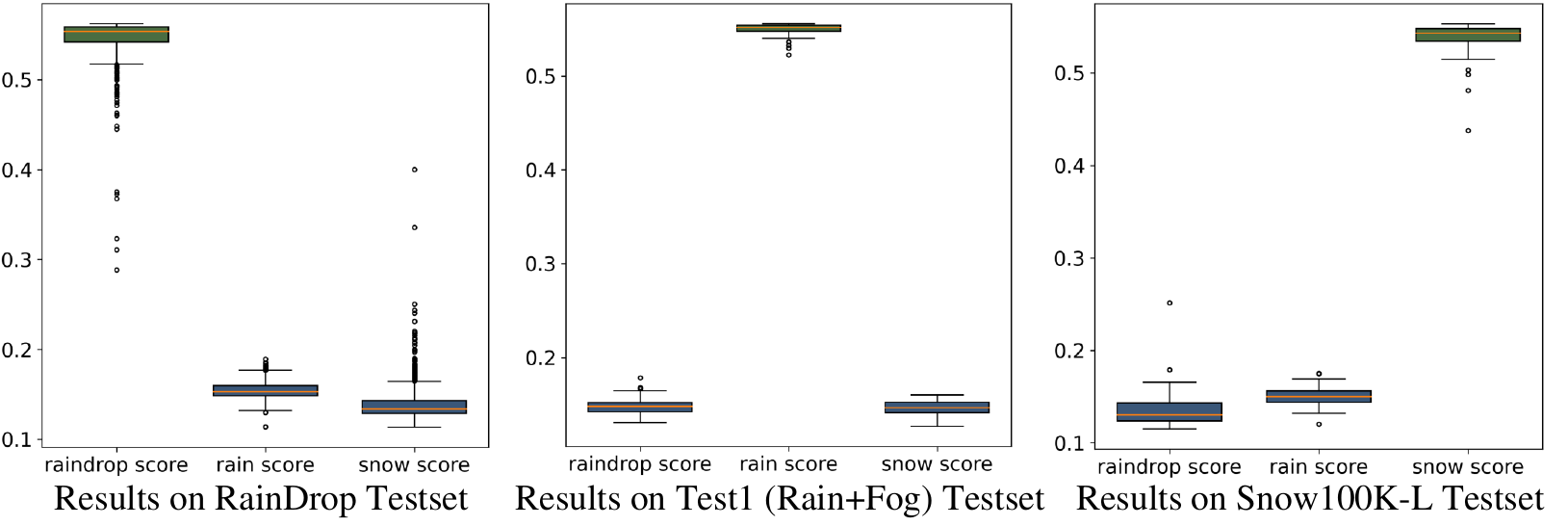}
    \vspace{-0.7cm}
    \caption{Boxplots of the weather scores of different datasets.}
    \vspace{-0.4cm}
    \label{boxplot}
\end{figure}

\begin{figure*}[!t]
\centering
\footnotesize
\def\yem{-3pt}
\def\xwidth{0.99}
\def\xxxwidth{0.162\textwidth}
\setlength{\tabcolsep}{1pt}
\begin{tabular}{C{\xxxwidth}C{\xxxwidth}C{\xxxwidth}C{\xxxwidth}C{\xxxwidth}C{\xxxwidth}}
\multicolumn{6}{c}{
\includegraphics[width=\xwidth\linewidth]{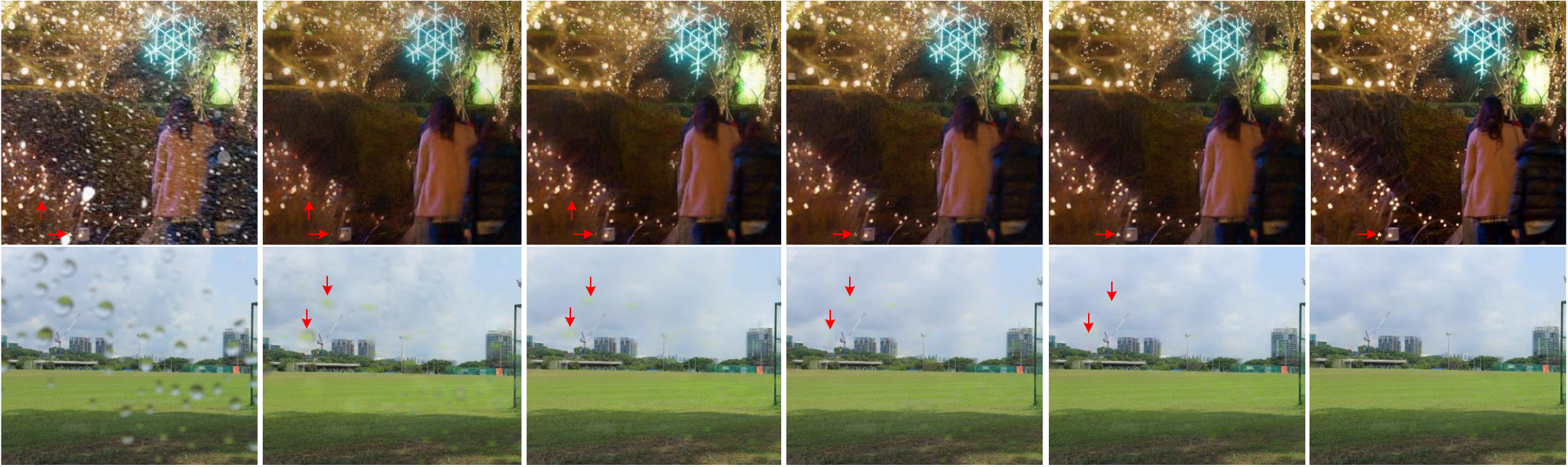}
} \\[\yem]
Input & baseline & +L & +L+G & +L+G+C & Ground Truth \\
\end{tabular}
\caption{Visualization of the ablation study. ``L", ``G" and ``C" denote local adaptivity, global adaptivity and channel-wise feature modulation respectively.}
\label{ablation_visualize}
\end{figure*}

We also visualize the results of the ablation study in Fig. \ref{ablation_visualize}. The baseline model (without the three proposed modules) cannot thoroughly remove the artifacts, as pointed out by the up arrow in the first row and the down arrow in the second row. Some image details are also treated as artifacts and thus blurred, as indicated by the right arrow. The image quality can be largely improved after adding the local adaptivity modules to the model, owing to the adaptive local operations. Then, by adding the global adaptivity module, the model gains a better global understanding of how to differentiate snowflakes or raindrops and their background. In the first row, the model treats the content pointed by the right arrow as a light bulb rather than a snowflake. In the second row, the artifacts with the same color as the grass are suppressed. Last, by adding the channel-wise modulation module, the image details are further enhanced, as pointed out by the right arrow in the first row.

\begin{figure}[t]
    \centering
    \includegraphics[width=1.0\linewidth]{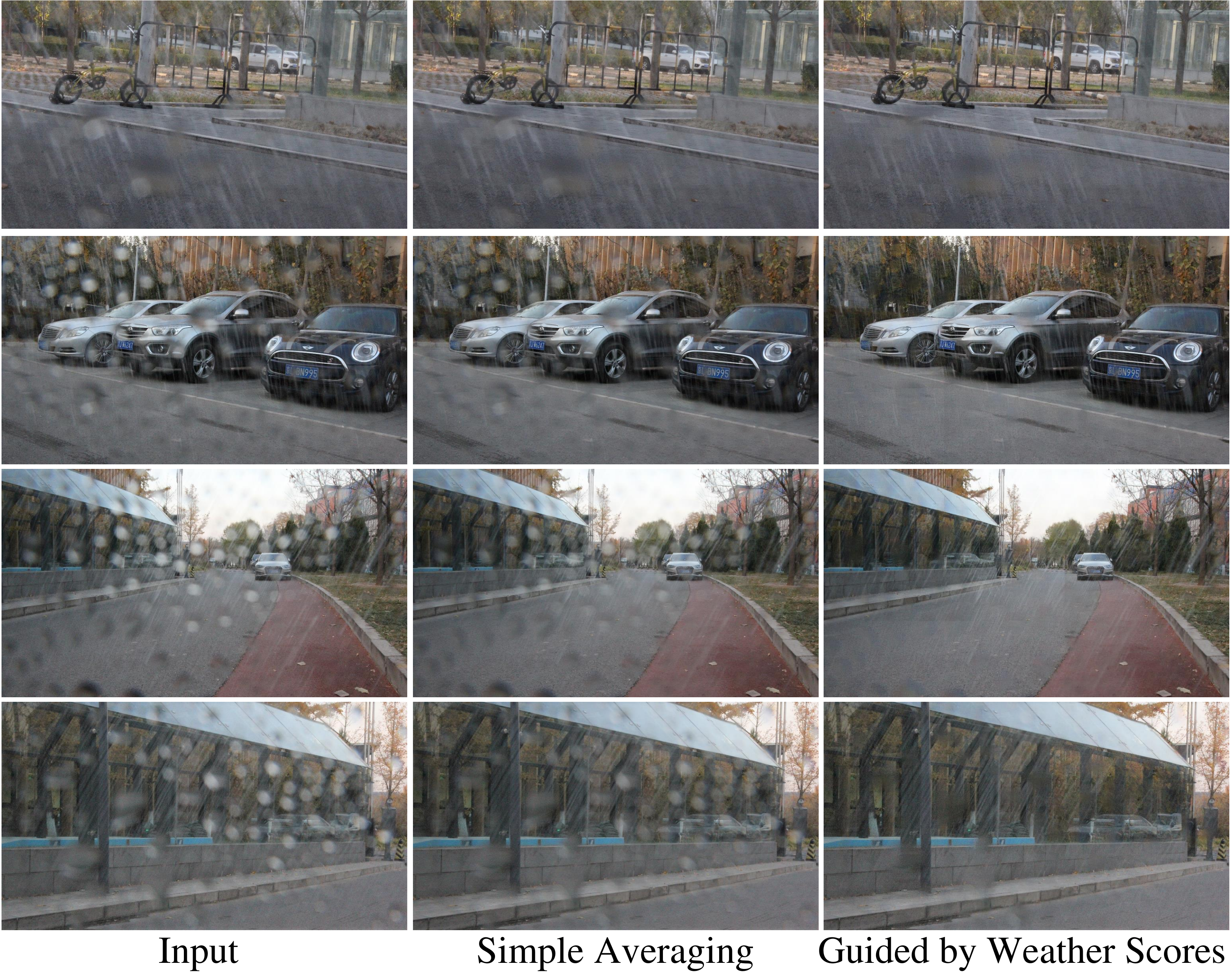}
    \vspace{-0.5cm}
    \caption{Comparisons between the simple averaging strategy and our weather-score-guided strategy on real-world hybrid-weather-degraded images \cite{RainDS}.}
    \vspace{-0.5cm}
    \label{two_strategies}
\end{figure}

\subsection{Results of Extended Applications}
The strategy for computing weather scores was tested on RainDrop testset \cite{Raindrop}, Test1 dataset \cite{OutdoorRain} (rain+fog), and Snow100K-L testset \cite{Snow100k}. Boxplots in Fig. \ref{boxplot} illustrate the distribution of weather scores for each dataset, showing that each dataset scored significantly higher for its corresponding weather type than for others. Of all 17,069 test images, only 2 were misclassified. Overall, our proposed weather score aligns with the type of weather existing in the picture.

We also tested the strategy for guiding pre-trained expert models on real-world images with hybrid degradations \cite{RainDS}. Three SOTA pre-trained models were selected as the weather-specific experts: AST \cite{AST} for raindrop removal, ConvIR-Rain \cite{ConvIR} for deraining, and ConvIR-Snow \cite{ConvIR} for desnowing. Due to the absence of high-quality ground truth, we present the visual results in Fig. \ref{two_strategies}. To simulate the possible scenarios in practical use, we also implemented a comparison strategy: processing input images with each expert model separately and then averaging the outputs. This approach reflects how systems without our hyper-network cannot determine the weather characteristics of the input and cannot select a suitable expert, leading to a simple fusion of results. As shown in Fig. \ref{two_strategies}, while the simple averaging strategy required more computation, their results were far from satisfactory. In contrast, using the proposed feature extraction hyper-network, we can compute the weather scores of the input image and accordingly select the most appropriate expert model to eliminate the most visually distracting degradations in the image.

\section{Discussions}
\label{discussions}

\subsection{Detailed Comparisons with Our Baseline}

\begin{figure}[t]
    \centering
    \includegraphics[width=1.0\linewidth]{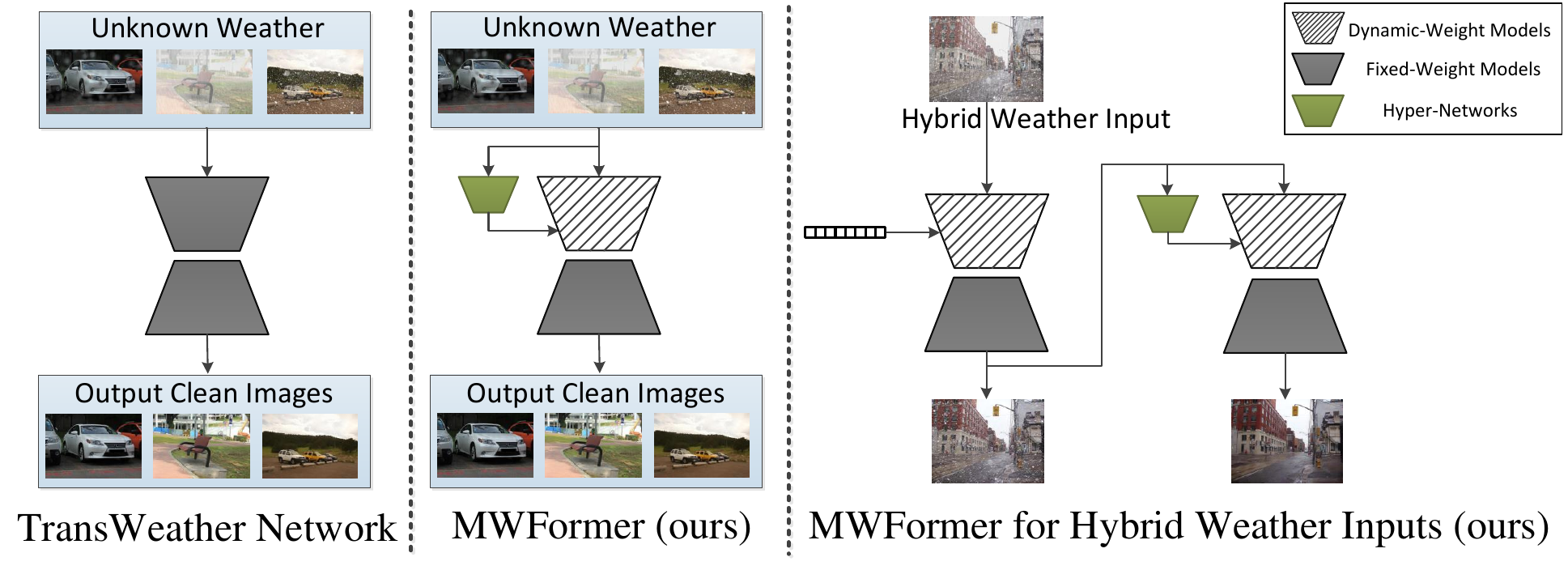}
    \vspace{-0.6cm}    
    \caption{Comparison between the baseline's architecture and our architectures.}
    \vspace{-0.4cm}
    \label{arch_compare}
\end{figure}

\textbf{Different architectures:} Fig. \ref{arch_compare} compares the architecture of MWFormer and our baseline model TransWeather \cite{TransWeather}.

As for the overall framework, TransWeather only contains an image restoration backbone, whereas MWFormer additionally uses a feature extraction network to guide the operations of the image restoration backbone adaptively. With a well-designed structure and training strategy, the feature extraction network extracts information related to weather features from the Gram matrices.

As for the architecture of the image restoration backbone, TransWeather employs a common image restoration network architecture with all parameters fixed, which lacks a special design for the task of multi-weather restoration. 
On the contrary, our model is specifically designed for multi-weather restoration by dividing the parameters into two groups, i.e., the fixed parameters encoding the general restoration knowledge, and the weather-adaptive parameters dynamically generated using the feature vector. In addition to operations in the parameter space, the feature vector also modulates the image restoration network in the feature space.

Additionally, two test-time variants have been developed: one for reducing the computational cost, and the other for handling hybrid adverse weather types unseen during training. The proposed MWFormer is the first model capable of restoring images degraded by the unseen hybrid adverse weather.

\textbf{Different applications:} The application of TransWeather is relatively limited, since it can only handle a few fixed weather types that have been already seen during the training phase. Our proposed MWFormer, with its flexibility, can restore the images impaired by \textbf{hybrid weather unseen during training}. This superiority over TransWeather indicates that MWFormer is more applicable to real-world scenarios, where different weather types may be commingled.
Moreover, the proposed feature extraction hyper-network not only can be combined with MWFormer's image restoration backbone, but also has a wider range of application scenarios, such as identifying the weather type, and guiding pre-trained weather-specific expert models, as introduced in Sec. \ref{extended_applications}.
Besides, we have also explored ways of training the image restoration model to benefit downstream detection tasks (Sec. \ref{task_driven_comparisons}), which is not addressed in TransWeather \cite{TransWeather}.

\begin{table*}[t]
    \centering
    \caption{Comparisons of performances of various network architectures with or without our proposed adaptive methodology. The performances of these network architectures can be significantly improved if combined with our methodology.}
    \begin{threeparttable}
    \centering
    \begin{tabular}{c|cc|cc|cc|cc}
    \toprule
        \multirow{2}{*}{Model} & \multicolumn{2}{c|}{RainDrop~\cite{Raindrop}} & \multicolumn{2}{c|}{Outdoor-Rain~\cite{OutdoorRain}} & \multicolumn{2}{c|}{Snow100K~\cite{Snow100k}}  &  \multicolumn{2}{c}{Average}  \\
               & PSNR $\uparrow$    &SSIM $\uparrow$   & PSNR $\uparrow$  & SSIM $\uparrow$   & PSNR $\uparrow$  & SSIM $\uparrow$  & PSNR $\uparrow$  & SSIM $\uparrow$  \\
    \midrule
    Restormer & 29.68 & 0.9042 & 28.29 & 0.8973 & 28.77 & 0.8768 & 28.91 & 0.8928\\
    \textbf{Ada-Restormer} & 29.80 & 0.9045 & 29.09 & 0.9035 & 29.21 & 0.8844 & 29.37 & 0.8975\\
    \midrule
    Uformer & 29.23 & 0.9266 & 25.41 & 0.8785 & 28.30 & 0.8732 & 27.65 & 0.8928\\
    \textbf{Ada-Uformer} & 29.88 & 0.9292 & 25.46 & 0.8841 & 28.82 & 0.8817 & 28.05 & 0.8983\\
    \midrule
    UNet & 29.19 & 0.9031 & 26.40 & 0.8857 & 28.60 & 0.8745 & 28.06 & 0.8878\\
    \textbf{Ada-UNet} & 29.70 & 0.9070 & 27.71 & 0.8975 & 29.11 & 0.8819 & 28.84 & 0.8955\\
    \bottomrule
    \end{tabular}
    \vspace{-0.2cm}
    \label{different_architectures}
\end{threeparttable}
\end{table*}

\subsection{Generalization Ability}
To demonstrate the generalization ability of our methodology, we integrated our approach into three different network architectures and evaluated the results: two Transformer-based architectures (Restormer \cite{Restormer} and Uformer \cite{Uformer}) and a CNN-based architecture (UNet \cite{UNet}). 

For each of the architectures mentioned above, we trained two versions of the model: one using the original network structure and the other combined with our proposed adaptive method (denoted as ``Ada-xxx"), both models with the same hyperparameters and number of channels. For Ada-Restormer and Ada-Uformer, we used the feature vector generated by the hyper-network to guide the restoration backbone across three dimensions and scales: locally spatial-wise, globally spatial-wise, and channel-wise. This allows part of the restoration backbone's parameters to be adaptively generated and its intermediate feature maps to be modulated based on the feature vector. Due to limited GPU memory, we reduced the encoder channels to 16 and 8 for the first scale of Ada-Restormer and Ada-Uformer, respectively, with a batch size of 16 for both. For Ada-UNet architecture, considering that CNN cannot capture long-range dependencies, we only applied the adaptivity locally spatial-wise and channel-wise. In addition, we removed the batch normalization layers in Ada-UNet and the original UNet architecture, which are commonly regarded as unsuitable for image restoration tasks. The other settings are the same as those reported in Sec. \ref{training_details}.

The quantitative results are reported in Table \ref{different_architectures}, indicating that our method can significantly improve the performance of \textbf{various network architectures} on multiple datasets. 
These promising results show that our proposed approach can be used as \textbf{a general approach} to boost the performance of different network architectures on multi-weather restoration tasks.

\begin{figure}[!t]
    \centering
    \includegraphics[width=0.65\linewidth]{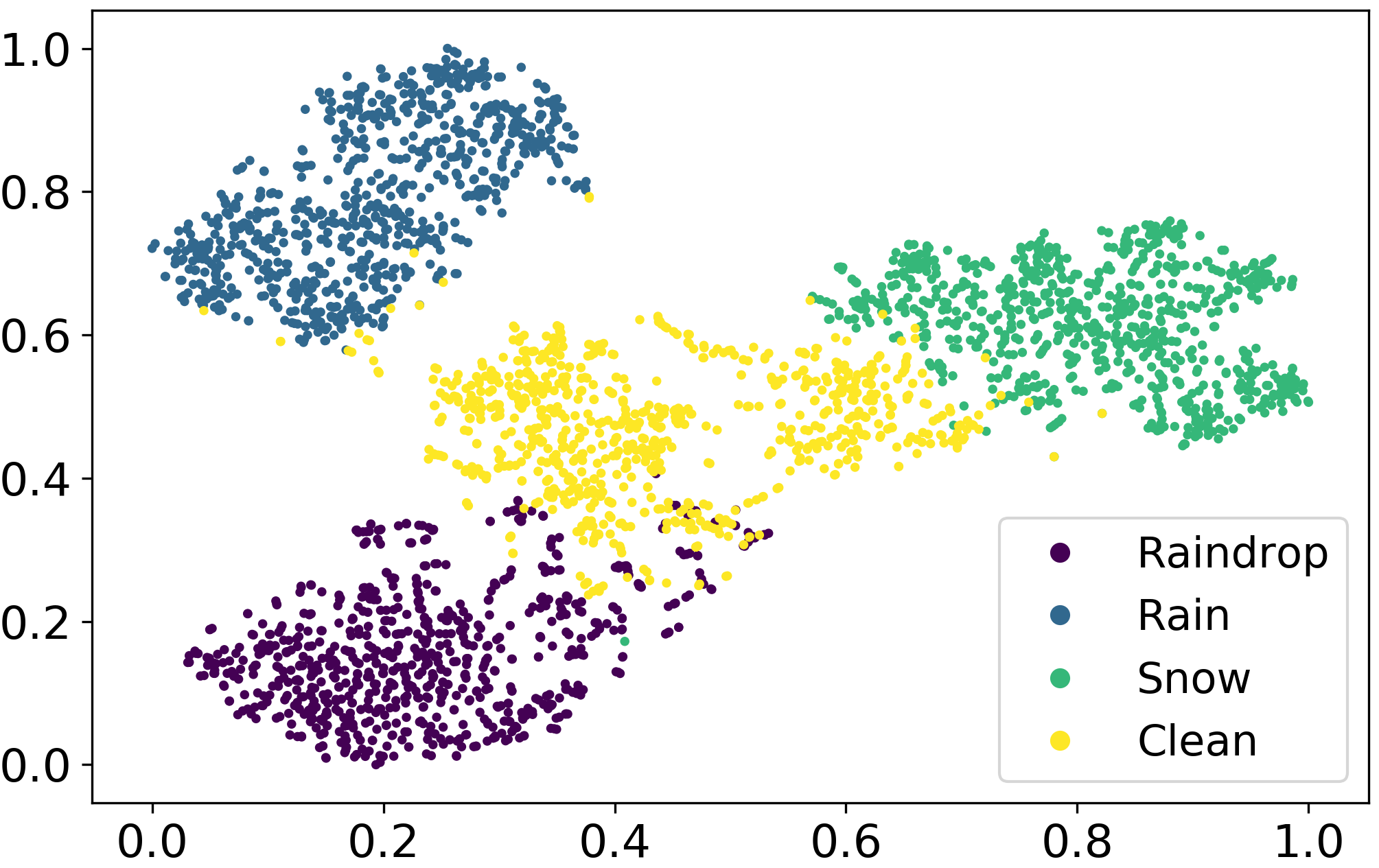}
    \vspace{-0.2cm}
    \caption{A t-SNE visualization of the distributions of the extracted feature vectors from different weather datasets. The feature extraction network learns effective embedding that is able to cluster images according to their weather degradation types.}
    \vspace{-0.4cm}
    \label{tsne}
\end{figure}
\subsection{Analysis of the Learned Weather Representation}
To better illustrate how the learned feature vector improves the performance of the image restoration network, we utilized t-SNE~\cite{van2008visualizing} to visualize the distributions of the weather-type features learned by the feature extraction network $\mathcal{F}_{feat}$.
As shown in~\cref{tsne}, the computed feature embeddings quite effectively decoupled the weather degradations accross contents, since images degraded by the same weather type become closely clustered with little overlap.
This suggests that the feature extraction network was able to learn to separate the content and degradation representations using contrastive loss.

We also examined the impact of feature vectors on image restoration using the simple version (\cref{variants}(b)) of MWFormer.
Using the raindrop removal as an example, we first tested the model on the Raindrop test set using the default setting for fixed weather degradation, meaning that the feature vector was the average of all the feature vectors of the raindrop images from the Raindrop training set.
We then replaced the default feature vector with the feature vector computed on an arbitrary image from the Raindrop testset and the Snow100K testset, respectively.
Numerical results in~\cref{simpler_version_exp} indicate that MWFormer performed the best when using the correct weather type embedding, demonstrating that average feature vectors effectively represent their corresponding weather types.
The performance slightly declined when using an arbitrary feature vector drawn from an image affected by the same weather type, and significantly dropped when using a feature vector of a different weather type.
{Generally, these results show that the vectors generated by our feature extraction network effectively encode weather-dependent information for guiding weather restoration tasks.
Finally, owing to the design of feature guidance of our MWFormer, the users have the capability to arbitrarily control the action of the image restoration network by providing a feature vector according to their prior knowledge.
This kind of flexibility during inference time is a key advantage that is unavailable in prior works.

\begin{table}[!t] 
\centering
\setlength{\tabcolsep}{6pt}
    \caption{Results of using different feature vectors in the simplified version of our model.}
    \begin{tabular}{cc|cc|cc}
    \toprule
    \multicolumn{2}{c|}{Average of Raindrop} & \multicolumn{2}{c|}{Arbitrary Raindrop} & \multicolumn{2}{c}{Arbitrary Snow} \\PSNR $\uparrow$    &SSIM $\uparrow$  & PSNR $\uparrow$ & SSIM $\uparrow$  & PSNR $\uparrow$ & SSIM $\uparrow$\\
    \midrule
    29.38 & 0.9073 & 26.93 & 0.8961 & 21.76 & 0.8139\\
    \bottomrule
    \end{tabular}
    \vspace{-0.4cm}
    \label{simpler_version_exp}
\end{table}

\section{Concluding Remarks}
\label{sec:conclusion}

We have introduced an efficient, all-in-one weather-aware Transformer, called MWFormer, for restoring images degraded by multiple adverse weather conditions.
%
MWFormer consists of an encoder-decoder-based restoration backbone, augmented by an auxiliary feature extraction hyper-network that learns weather-type representations.
%
The extracted feature vectors can be used to adaptively guide the main image restoration backbone by weight-adaptivity along the local, global, and channel axes.
They can also be used for weather-type identification or guiding pre-trained expert models.
Because of the availability of the auxiliary network, MWFormer can be extended to deal with fixed single-weather cases with less computation or hybrid-weather cases that were unseen during training.
We conducted a spectrum of quantitative and qualitative studies on the multi-weather restoration benchmark dataset as well as on real-world datasets, and the results show that MWFormer outperforms prior known multi-weather restoration models without requiring much computational effort.
Our methodology can also be integrated into a variety of network architectures to boost their performance.


\bibliographystyle{IEEEtran}

\normalem
\bibliography{ref}

\end{document}